\documentclass[a4paper,fleqn]{cas-dc}

\usepackage[numbers]{natbib}

\usepackage{booktabs}
\usepackage{siunitx}
\usepackage{multirow}
\usepackage[table]{xcolor} 
\usepackage{stfloats}
\usepackage{tabularx}
\usepackage{cleveref}
\usepackage{placeins}
\usepackage{makecell}
\usepackage{pifont}

\definecolor{bestyellow}{HTML}{FFF2CC}  
\definecolor{secondpurple}{HTML}{EAD1DC} 

\newcommand{\best}[1]{\colorbox{bestyellow}{#1}}
\newcommand{\secondbest}[1]{\colorbox{secondpurple}{#1}}
\newcommand{\y}{\textcolor{green!55!black}{\ding{51}}}
\newcommand{\n}{\textcolor{red!70!black}{\ding{55}}}
\newcommand{\na}{\textcolor{gray}{--}}
\newcommand{\hf}[1]{\href{https://huggingface.co/#1}{\texttt{#1}}}
\newcommand{\gh}[1]{\href{https://github.com/#1}{\texttt{#1}}}

\begin{document}
\let\WriteBookmarks\relax
\def\floatpagepagefraction{1}
\def\textpagefraction{.001}

\shorttitle{MDTD-ArtIR}    

\shortauthors{}  

\title [mode = title]{MDTD-ArtIR: Benchmarking Image Editing and Restoration Models for Art Image Restoration under Texture-Overlay Degradations}  



%

\author{Mridula Vijendran}[orcid=0000-0002-4970-7723]
\ead{mridula.vijendran@durham.ac.uk}
\credit{Conceptualization, Methodology, Data curation, Validation, Visualization, Writing – original draft}

\affiliation{
    organization={Department of Computer Science, Durham University},
    city={Durham},
    country={United Kingdom}
}

\author{Shuang Chen}[orcid=0000-0002-6879-7285]
\ead{shuang.chen@durham.ac.uk}
\credit{Conceptualization, Methodology, Data curation, Writing – review and editing}

\author{Hubert P. H. Shum}[orcid=0000-0001-5651-6039]
\cormark[1]
\ead{hubert.shum@durham.ac.uk}
\credit{Conceptualization, Writing – review and editing, Supervision}

\cortext[1]{Corresponding author}



\begin{abstract}
Restoring severely degraded visual media still remains a formidable challenge, as existing methods often hallucinate unnatural textures and contents, struggle with preserving color and texture, or fail to leverage partially retained image information.  Existing restoration benchmarks assume known degradation operators and fail to capture the complex characteristics of artistic damage such as cracks, stains, and color/texture deviation. We introduce a controlled benchmark for blind restoration of semantic, semi-transparent image media degradations, accompanied by a new degradation alpha texture mask dataset MDTD-Art. We present a new dataset and benchmark evaluating state-of-the-art universal restoration models against image editing and vision-language models across varying mask opacity levels. Our experiments demonstrate that image editing models consistently outperform specialized restoration architectures for arbitrary degradations, with performance gains amplified by structured prompt engineering emphasizing detail preservation and structural consistency. These findings position recoverable semantic information and prompt controllability as critical factors in art image restoration.
\end{abstract}





\maketitle

\section{Introduction}
\label{sec:intro}

\begin{figure*}[t]
    \centering

    \begin{tabular}{cccccc}
        \textbf{Degraded Image} & \textbf{GPT Image 1.5} & \textbf{NB Pro} & \textbf{Flux 2} & \textbf{AutoDIR}& \textbf{Ground Truth} \\


        \includegraphics[width=0.12\textwidth]{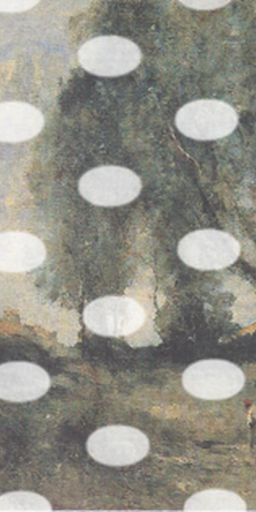} &
        \includegraphics[width=0.12\textwidth]{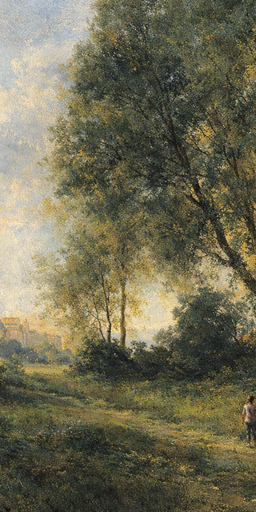} &
        \includegraphics[width=0.12\textwidth]{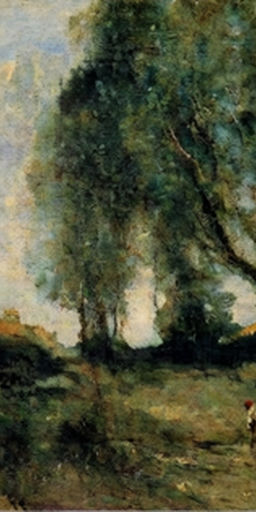} &
        \includegraphics[width=0.12\textwidth]{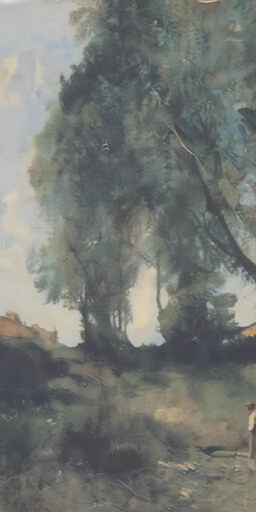} &
        \includegraphics[width=0.12\textwidth]{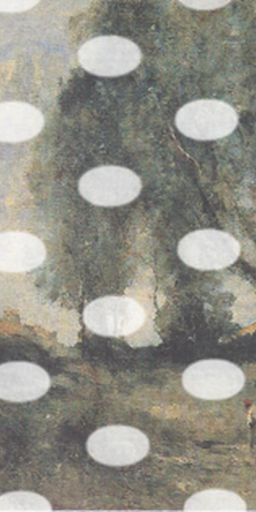} &
        \includegraphics[width=0.12\textwidth]{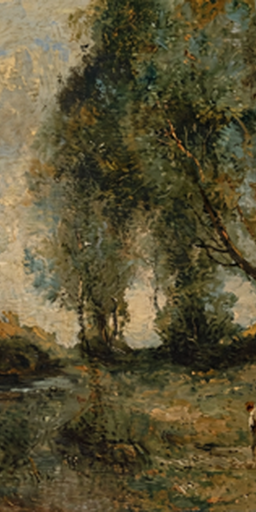} 
        \\ \vspace{0.5ex}

        \includegraphics[width=0.12\textwidth]{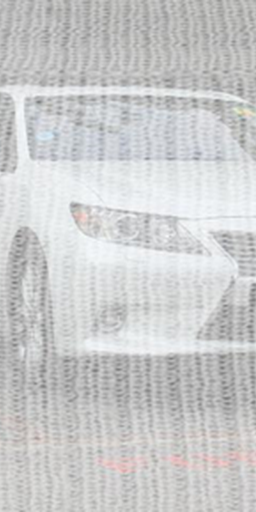} &
        \includegraphics[width=0.12\textwidth]{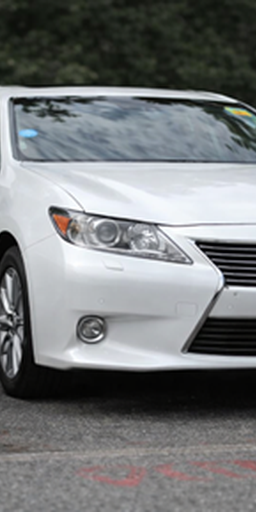} &
        \includegraphics[width=0.12\textwidth]{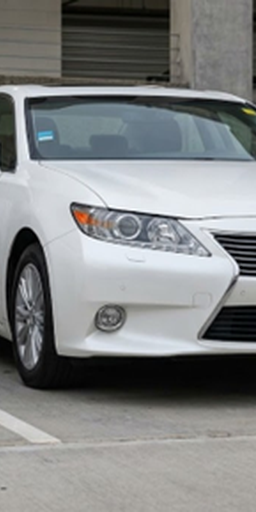} &
        \includegraphics[width=0.12\textwidth]{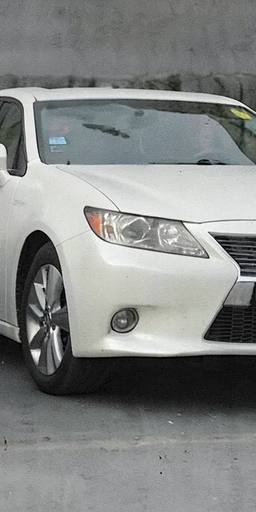} &
        \includegraphics[width=0.12\textwidth]{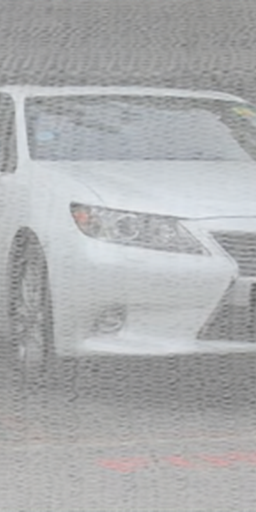} &
        \includegraphics[width=0.12\textwidth]{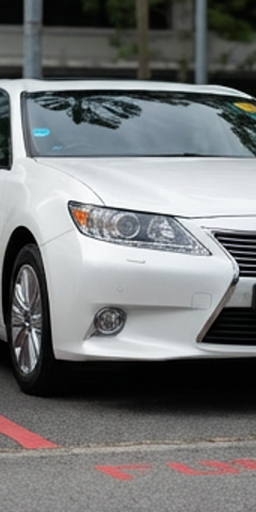} \\ \vspace{0.5ex}

        \includegraphics[width=0.12\textwidth]{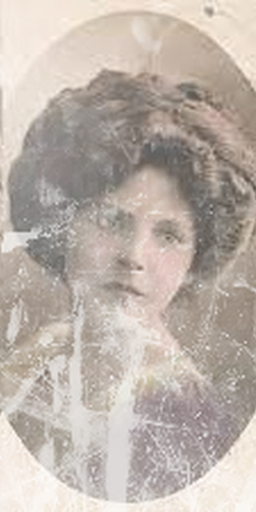} &
        \includegraphics[width=0.12\textwidth]{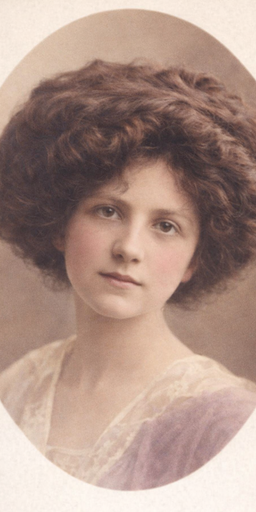} &
        \includegraphics[width=0.12\textwidth]{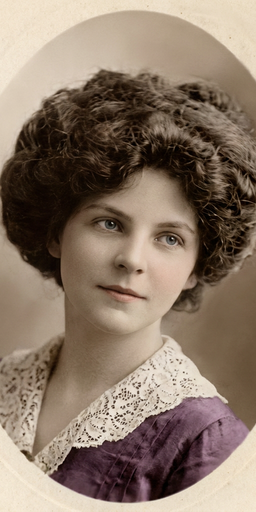} &
        \includegraphics[width=0.12\textwidth]{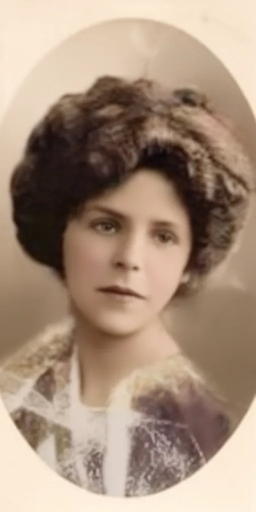} &
        \includegraphics[width=0.12\textwidth]{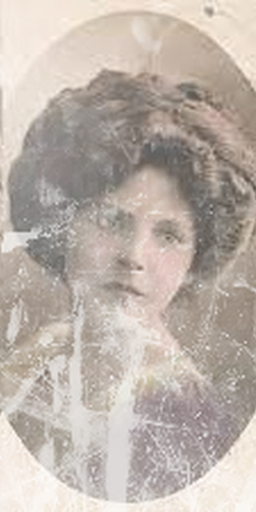} &
        \includegraphics[width=0.12\textwidth]{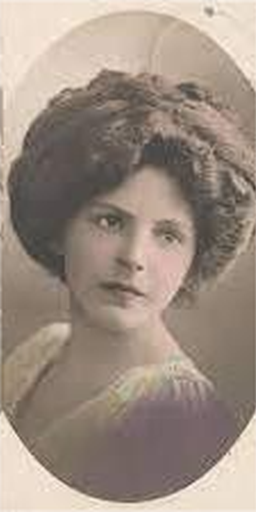} \\ \vspace{0.5ex}

        \includegraphics[width=0.12\textwidth]{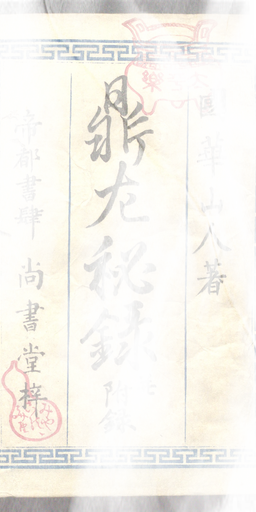} &
        \includegraphics[width=0.12\textwidth]{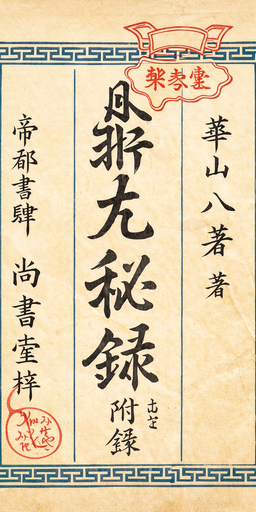} &
        \includegraphics[width=0.12\textwidth]{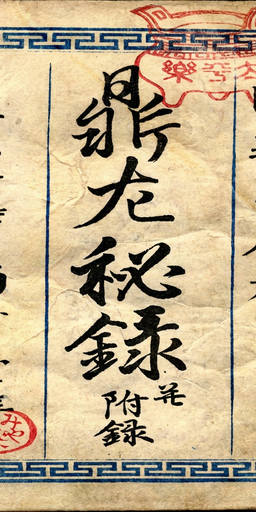} &
        \includegraphics[width=0.12\textwidth]{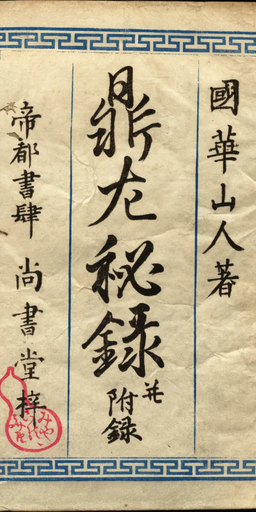} &
        \includegraphics[width=0.12\textwidth]{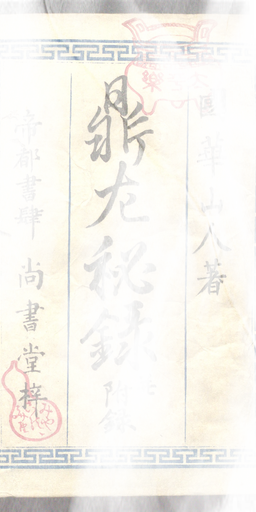} &
        \includegraphics[width=0.12\textwidth]{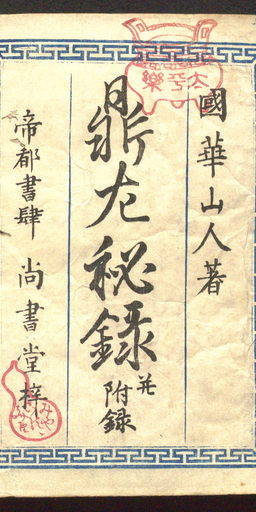} \\ \vspace{0.5ex}
    \end{tabular}

    \caption{Visual comparison of image restoration results on four sample images from different domains with cropped native model outputs for same view comparisons. The Masked image is shown in the first column, followed by predictions from models GPT Image 1.5, Nano Banana Pro, Flux 2 and AutoDIR, and finally shows the ground truth .}
    \label{fig:visual_comparison_grid}
\end{figure*}

AI-driven art image restoration aims to recover degraded content while preserving visual fidelity and perceptual coherence through colorization, inpainting, de-cracking, and stain removal. These models aid preservation efforts for historical photographs \cite{wan2020bringing, Luo-Rephotography-2021}, paintings \cite{garcia2025artinsight} and manuscripts \cite{cai2024hierarchical, shah2020performance} that have deteriorated due to aging or damage from the environment or preservation efforts. Datasets for damaged artistic mediums are limited in scope \cite{xu2024comprehensive, yang2023hq}, with the domain consisting of a small number of degraded samples like wall paintings. They include paired and unpaired collections of unidentified damaged paintings over degradation classes categorized data, or cover only synthetic mask overlays or photorealistic media common degradation operations. Additionally, SOTA image restoration models are not trained in art-restoration tasks \cite{yin2026far} but image editing models \cite{sun2026can} that do so are not optimized to maintain fidelity at the pixel and perceptual level.

Existing work covers degradation operations that are known such as weather degradations and low-light conditions, as well as digital degradations including compression, noise distributions, and blur transformations \cite{yin2026far} for both real-world images and non-photorealistic media. Other parts of the generated images commonly have unnatural textures, hallucinated details, or color gradient alterations from repeated inpainting and colorization \cite{o2023limitations}. Conditional mask-guided methods, even when combined with text embeddings for texture and color context, frequently fail to capture fine-grained details of abstract/unique contents \cite{huang2024learning} and blend them with other visual elements \cite{vinker2023concept}, resulting in noticeable inconsistencies in subject identity and appearance. Moreover, Blending-based methods in embedding space \cite{vinker2023concept} or at the diffusion level \cite{zheng2025lanpaint, chang26design} that provide the model with more freedom to integrate diverse content of the same modality fail to maintain local coherence or produce boundary artifacts. Prompt-engineering in VLMs, a technique that guides model behavior on inputs like images for tasks such as image retrieval \cite{sah2025retrieval}, has proven effective in other areas such as targeted image editing \cite{yilmaz2026edit2restore} or restoration \cite{bai2025textir}. The models that utilize these methods can be broadly categorized into vision language models (VLM), image editing models that re-purpose VLMs by working on top of an input image with conditional prompt instruction or mask information, and universal image restoration (UIR) models that combine the above methods and different architectures on the specific task of image restoration but are applicable for a large number of domains and degradation operators.

Restoration of damaged visual media remains challenging due to the heterogeneous nature of degradation. Damaged regions may partially retain the surrounding image information, such as faded colors, blurred details from aging, or yellowed areas, or they may completely lose information through cracks, peeling, or material loss. These degradations are typically caused by environmental factors such as light exposure, heat, moisture, mold, poor storage conditions, or aggressive chemical treatments. Existing methods struggle to address this variability: some assume damaged regions contain no usable information, limiting their ability to leverage partial cues \cite{avrahami2023break, safaee2024clic}; others rely on prior knowledge about the location or extent of damage \cite{wang2025towards}; and many focus primarily on global noise effects such as speckle, salt-and-pepper noise, lossy compression, or Gaussian blurring \cite{yang2023pasd, guo2025mambair}, which do not capture the complex, localized characteristics of real-world degradation. Thus, exploring synthetic datasets provides value to bridge this gap, as seen with other domains in the same restoration task \cite{shamsuddin2022synthetic}.

Prompt design together with large image editing models has previously demonstrated generalizable image restoration across different degradation types and scenes \cite{zuo2025nano, yang2026realrestorer, sun2026can} and improved the perceptual quality of images \cite{you2025enhancing}. While they can match the performance of specialized model, they are highly sensitive to prompt design, progressive restoration, and provide stochastic performance, needing improvements in their controllability, stability, and robustness. In the worst-case scenarios, they lead to hallucinations, semantic and identity alterations, as shown in \cref{fig:hard_samples}, where some identity changes include human faces, scene changes and pattern hijacking. We reduce the visibility of the underlying scene at various levels of severity by overlaying semantically varied textured masks from the DTD dataset \cite{cimpoi14describing} on the image. This is similar to a weather degradation task in image restoration, but also provides a textured mask that needs to be inpainted inside, which is a task common in image editing models forming a hybrid restoration setting \cite{chen24hint}. The overlap of tasks justifies our use of large image editing and image restoration models. Additionally, the textures form a synthetic abstraction of stains and cracks common in damaged artworks. While our degradations are not a substitute for physical simulations, we observe a similarity in our qualitative experiments where the masked image is similar to the real-world degradations, along with the model generating outputs with similar observed characteristics. To this end, we construct a alpha texture mask dataset MDTD-Art (Mask Describable Textures Dataset for Artworks media) that approximates natural real-world degradation patterns (such as cracks, stains, and material loss) through combinatorially generated alpha-textured overlays, introducing degradation classes and severities that are not covered by existing degradation-specific datasets or restoration models specialized to known operators, thereby defining a controlled synthetic stress test for restoration under various levels of semantic texture corruptions. To address all the gaps in the existing works, we introduce a dataset (MDTD-Art) to generate various semantic texture corruptions, a hybrid restoration-inpainting task (MDTD-ArtIR) for recovering damaged regions while utilizing their underlying semantic information, and a benchmark evaluating recent image editing and restoration models under varying opacity levels across Image Quality Assessment (IQA) metrics.

To showcase the difficulty of the task we call MDTD-Art Image Restoration, we look at different model restorations of specialized universal restorers against our image editing models in the experimental results section, along with some hard examples where the models restored the image following the mask's textural cue, forgoing the identity of the underlying image. The selected image-editing models perform strongly at lower opacity but demonstrate failure under ambiguous texture patterns or higher opacity levels. Furthermore, a quantitative comparison across varying degrees of mask opacity reveals that explicit degradation-aware instructions improve the models, provided the model has the bandwidth for both the input image and prompt tokens, emphasizing detail preservation and structural consistency leads to more faithful reconstructions (especially in Nano Banana (NB) Pro) for other types of masks. Our task difficulty can be visualized in Figure \cref{fig:visual_comparison_grid} to check for the reconstruction of fine colors, intricate textures, and sharp semantic details even under the most dense degradation overlays. We observe that image restoration-specific models (e.g. AutoDIR \cite{jiang2024autodir}) are not able to handle the various degradation types from our dataset as well as the large image editing models (e.g. GPT Image 1.5 \cite{openai2025gptimage15}, NB Pro \cite{sun2026can,team2023gemini}, Flux \cite{labs2025flux}). The image editing models tend to change the semantic details or colors. They may also change the aspect ratio of the input image.

Our main contributions are two-fold. We construct and release MDTD-Art, an alpha texture mask dataset that combinatorially pairs the Describable Textures Dataset (DTD) with clean art images to form degradations at controlled alpha-opacity levels. The construction allows texture class and an opacity-based severity parameter to be varied independently, for different domains, and in a combinatorially scalable way \cref{sec:mtao} \cref{fig:dtd_alpha_mask}. We further conduct a benchmark study on recent general-purpose image editing models showing the effectiveness of restoration by generic to explicit degradation-aware prompt-guided image restoration.

\section{Related Works}
\label{sec:formatting}

\subsection{Image Editing Models}
Recent image editing systems have shifted from task-specific pipelines to general-purpose editing foundation models that can handle multiple instructions, generalizing to various tasks and being trained on large image-text pairs and often performing better than specialized restoration models. A major design goal in this family is fidelity preservation, which retains subject identity, spatial structure, or simple/mono-lingual typography from the input image to the output image while only changing the requested attributes.

Closed image editing models with agentic features such as Nano Banana (NB) Pro  \cite{sun2026can} followed by GPT Image 1.5\cite{openai2025gptimage15} had previously achieved the best restoration quality and content consistency \cite{yang2026realrestorer}. These support prompt refinement through reasoning models and LLMs, reference and grounding images through search engine tools, and masks for controlled editing.

On the other hand, open source models such as Flux 2 \cite{labs2025flux}, Qwen Image Edit 2511 \cite{wu2025qwen}, FireRed \cite{team2026firered}, or Step1X-Edit \cite{liu2025step1x} offer more flexibility in use through their accessible model weights, opening techniques such as LoRA finetuning, external encoders, auxiliary adapters, and post hoc fusion for adapting to different tasks. They suffer from more involved user steering, lack of controlled outputs, and limited training data due to users' model usage.

\subsection{Universal Image Restoration}
UIR models handle any unknown, mixed, or singular source of degradation found in images of varying degree of severity found at different spatial opacities. These are broadly organized around two paradigms: those that rely on explicit degradation information to condition restoration, and those that extract degradation representations implicitly from the corrupted input itself.

A prominent line of work incorporates task-specific modules, embeddings, or priors as explicit conditioning signals to guide the restoration process. DiffBIR \cite{lin2024diffbir} introduces task-specific restoration modules as diffusion priors, leveraging a ControlNet architecture in a two-stage pipeline that first applies a task-conditioned restoration module before refining the output via diffusion. Similarly, BIRD \cite{chihaoui2024blind} uses the output of the task-based model as explicit degradation priors to initialize a fast diffusion inversion process, reducing computational overhead while preserving the expressiveness of the diffusion-based priors. SUPIR \cite{yu2024scaling} adopts a complementary approach, employing a degradation-robust encoder alongside ControlNet conditioning to inject degradation-aware signals into the diffusion prior. AdaptBIR \cite{liu2024adaptbir} takes a more architectural stance, introducing task-specific CNN encoders and decoders that extract multi-level degradation features from multiple latent layers, enabling the model to handle hybrid and multi-level degradations simultaneously. DA-RCOT \cite{tang2025degradation} approaches explicit conditioning through optimal transport theory, constructing residual embeddings that encode both degradation type and severity as structured conditioning inputs to a Restormer backbone. X-Restormer \cite{chen2024comparative} similarly employs task-specific backbone networks, maintaining separate restoration pathways per degradation class while sharing representational capacity across tasks. OneRestore \cite{guo2024onerestore} combines a degraded image with a scene description embedding as joint inputs to a transformer-based encoder-decoder architecture, using semantic scene context as a soft degradation prior.

Alternatively, existing universal restorers learn  to extract degradation-relevant features directly from the corrupted observation. BIR-Adapter \cite{eteke2026bir} exemplifies this approach, implicitly extracting degradation features from the latent representation of the corrupted image through self-attention refinement, without requiring any explicit degradation conditioning. RealRestorer \cite{yang2026realrestorer} bridges the gap between synthetic and real-world degradations through a two-stage progressive training strategy: an initial stage maps corrupted synthetic images to clean outputs using a weak degradation-like objective, followed by supervised fine-tuning on real corrupted-to-clean pairs. This progressive mixing curriculum improves generalization to the complex, uncharacterized degradations present in real-world imagery. 

\subsection{Prompt-Based Restoration}

A growing line of work explores natural language and learned prompt representations as conditioning modalities for universal restoration, enabling flexible degradation specification without requiring hard categorical degradation labels at inference time. These methods broadly divide into two paradigms: those that augment the restoration network with additional learned modules to bridge the text-image gap, and those that encode degradation context entirely within learned prompt representations, leaving the base restoration network unmodified.

Several works introduce dedicated auxiliary modules to align textual and visual representations for restoration guidance. Learnable prompts provided better degradation-aware alignment, leading to better restoration quality, albeit in limited weather-damaged media \cite{monga2025dairnet, jin2025degradation}. Edit2Restore \cite{yilmaz2026edit2restore} employs task-specific text prompts alongside a LoRA adapter and fine-tuned text encoder, steering a generative model toward degradation-specific corrections without full model retraining. TextualDegRemoval \cite{lin2024improving} introduces a CLIP-based image-to-text mapper paired with a dedicated text restoration module, injecting multi-scale visual information into the restoration network by operating directly on the textual embedding space, thereby removing degradation representations at the semantic level before propagating corrections to pixels. Res-Captioner \cite{sun2024beyond} takes a linguistically explicit approach. It adds a chain-of-thought captioning module that encodes both the degraded image and the degradation information into a structured natural language caption via an image-to-text model, which then guides the downstream restoration network. TextIR \cite{bai2025textir} similarly conditions restoration on CLIP-encoded text-image embedding pairs for degraded and ground truth images. Other works have explored prompts as alternatives for providing alpha texture masks \cite{vijendran25artificial} to intuitively describe editable regions \cite{fang2026region}.

\section{Methodology}
\begin{figure*}[]
    \centering
    \includegraphics[width=0.9\linewidth]{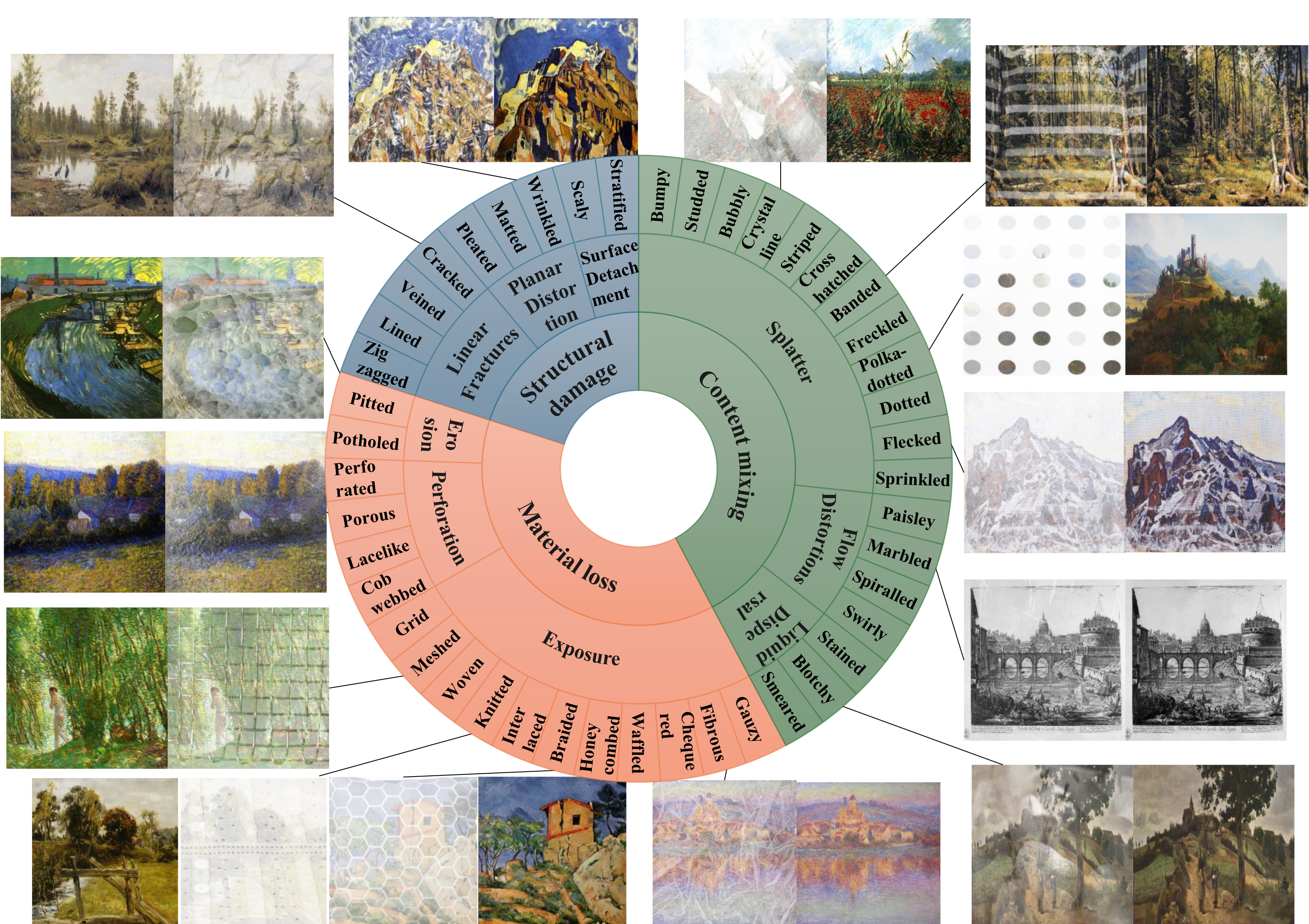}
    \caption{Overview of Data Distribution as a heuristic visual grouping. The DTD dataset is selected to approximate different types of damages found in artworks and manuscripts and further stratified into levels of mask opacities. The mask images are class-wise categorized into 3 levels of alpha groups based on the average alpha channel values.}
    \label{fig:dtd_alpha_mask}
\end{figure*}

\subsection{The MDTD-Art Dataset}
Real-world artistic damage (cracking, flaking, staining, material loss etc.) is characterized by pigment-dependent or local corruptions that are missing in existing restoration benchmarks. We select texture classes from the Describable Textures Dataset (DTD) \cite{cimpoi14describing} as a proxy for this structure such as cracked, flecked, stained, and porous \cref{fig:dtd_alpha_mask} share the same irregular, locally correlated spatial statistics as naturally occurring artwork damage, while remaining fully decoupled from any specific damaged-image source thereby allowing us to composite them onto any clean image domain rather than being tied to a fixed set of already-damaged artifacts, as in prior datasets \cref{tab:dataset_comparison}. These include HQ-50K \cite{yang2023hq50k} for open-world image restoration models with synthetic non-photorealistic image degradations and ArtInsight \cite{chheda2025artinsight} that uses the base dataset ArtDet for natural paint damage. Additionally, in the table, we compare MuralDH's \cite{xu2024comprehensive} 1000 Dunhuang subset statistics for wall painting natural damages against ours, which achieves synthetic degraded images that can be combinatorially generated with any domain dataset to get the corresponding texture degradation. 

Additionally, the alpha-transparency formulation \cref{eq:transparency_mask} lets us control information loss severity independently of degradation type rather than each texture class, implying a fixed corruption strength as in fixed-operator degradation datasets. We do a mean aggregate of alpha opacity of every class across low ($[0,0.33)$), medium ($(0.33,0.66]$), and high ($(0.66,1]$) levels. This produces a degradation regime that spans a wider range of information loss from mask opaqueness than what we qualitatively observe in the naturally aged artworks used in our real-world comparison \cref{fig:realworld_samples}, while the lower end of this range remains representative of typical aging patterns. By exceeding the severity range of natural damage, we stress-test whether restoration models rely on genuine semantic and structural priors versus shallow interpolation from lightly corrupted regions.

This construction also yields degradation classes that are, to our knowledge, not represented in existing degradation-specific restoration literature. Prior work targets known, parameterized operators (Gaussian blur, compression artifacts, weather effects, etc.) with well-defined inverse models; our texture-mask overlays instead define a fixed set of 47 texture corruption classes (\cref{tab:dataset_comparison}) with no known closed-form inverse. Since we hide the true degradation mask at inference, it closely mirrors the ill-posed nature of real damage. Consequently, image restoration models trained on or specialized for fixed degradation operators have no direct prior for this degradation family, which we show in \cref{sec:quant_res} and  \cref{sec:qual_res}, contributing to a lower performance relative to general-purpose image editing models, barring other model differences.

\subsection{Mask Texture Alpha Overlay}
\label{sec:mtao}
Our task involves recovering the media from arbitrary texture damage overlaid on them. We simplify the problem as an alpha layer overlay on the image, creating varying levels of information loss from the target image. 

This makes the goal of the model to separate which proportion of the image part belongs to the image itself or the damage, especially when the mask texture can be ambiguous with the art painting or photo's texture. This alpha overlay is described as:
\begin{equation}
\label{eq:transparency_mask}
\mathbf{I}_{deg} = \alpha \odot \mathbf{1} + (1 - \alpha) \odot \mathbf{I}_{clean},
\end{equation}
where $\alpha \in [0,1]^{H\times W} $ represents the grayscale alpha texture mask to overlay upon the clean image's ($\mathbf{I}_{clean}$) alpha channel to get the degraded image $\mathbf{I}_{deg}$. Our novel data augmentation strategy involves the pairing of each image with multiple texture classes of masks to increase variability in the corruption patterns. To prevent class bias of textures seen during training of the auxiliary mask-prediction model (UNet EfficientNet-B0), the masks are grouped and filtered with DTD unshuffled dataset's classes which are randomized at the start of each epoch. Let $C$ denote the number of mask classes and $S$ represents the number of mask class samples subset. To prevent bias of some classes over the other, the dataset length is defined as $N \times (CS)$ where N is the number of clean images to be masked and we have different mask overlays per clean sample. This allows for class-wise stratified sampling of data or allow for more intra-class variance. For a zero indexed dataset index $i$, the corresponding mask for the index $m$ is computed as: 
\begin{equation}
m = i \bmod (CS),
\end{equation}
and $i \in [0,N \times (CS))$ 
The subset indices for the mask class $m_i$ is determined by:
\begin{equation}
    m_i = \left\lfloor \frac{m}{C} \right\rfloor,
\end{equation}
where $m_i \in [0,S)$.

When training the mask predictor backbones in \cref{tab:unet_backbone_comparison}, we select $C = 47$ for the DTD class variations with $S=50$ randomized samples subset, but when doing model benchmarking experiments, our $C = 3$ for the low, medium and high opacity masks and $S$ selected based on samples in the classes that fit in the $C$ alpha thresholds limits.

This data augmentation strategy, along with image transformations such as horizontal flips, rotates for the input and target images allow the model to learn with small image datasets and handle mask-conditioned invariance during the VLM and image restoration model inference, and learn better mask-image correspondence. This is particularly challenging since our datasets involve a texture specific to the artwork, which can be ambiguous with translucent overlays from the mask. This allows the model to see many ways of breaking and recombining the mask and image rather than memorizing a few canonical patterns. 

\begin{table*}[t]
\centering
\caption{\textbf{Comparison of Image Restoration Datasets.} We compare our proposed MDTD-Art against prominent existing benchmarks across degradation diversity, image domain, and scale.}
\label{tab:dataset_comparison}
\footnotesize
\begin{tabularx}{.8\linewidth}{l X c l r}
\toprule
\textbf{Dataset} & \textbf{Degradation Types} & \textbf{\# Deg.} & \textbf{Clean Image Domain} & \textbf{\# Images} \\
\midrule
HQ-50K \cite{yang2023hq50k} & Rain, Compression artifacts, Noise, SR & 4 & Photos, Text, Maps, Po. & 50,000 \\
MuralDH \cite{xu2024comprehensive} & SR, Cracks, Flaking, Fading & 4 & Wall Paintings & 1,000 \\
ArtInsight \cite{chheda2025artinsight} & Lacune Material Loss & 1 & Paintings & 30 \\
\textbf{MDTD-Art} (Ours) & DTD texture alpha mask & 47 & Paintings & 5,640 \\
\bottomrule
\end{tabularx}
\end{table*}

\subsection{Compared Methods}

\begin{table*}[t]\centering
\caption{Configuration of the evaluated models. All models take the textured/masked image as
input; text (prompt) input follows the generic/explicit prompt design
(\cref{tab:prompt_comparison}); mask-argument models receive a proxy mask from the
EfficientNet-B0 UNet in the final row. Every output is Lanczos-resized so its longest side is
512\,px and deterministically center-cropped to $512{\times}512$. Open-weight models were run
with \texttt{diffusers}~\texttt{X.Y.Z} and \texttt{torch}~\texttt{X.Y.Z} between
\textit{[start]} and \textit{[end]}. Hugging Face identifiers resolve under
\texttt{huggingface.co/}, GitHub identifiers under \texttt{github.com/}; hexadecimal entries in
the last column are the exact commits used.}
\label{tab:model_config}
\footnotesize
\renewcommand{\arraystretch}{1.25}
\begin{tabular}{l c c l c l}
\toprule
\textbf{Model} & \makecell{\textbf{Text} \textbf{In}} & \makecell{\textbf{Mask} \textbf{In}} & \textbf{Repository / API identifier} & \textbf{Precision} & \textbf{Version/Revision} \\
\midrule
\multicolumn{6}{l}{\cellcolor{gray!12}\textit{Closed image-editing (agentic VLM) - proprietary API}} \\
GPT Image 1.5   & \y & \n & \texttt{gpt-image-1.5} (OpenAI) & \na & 16 Dec 2025 \\
Nano Banana Pro & \y & \n & \texttt{gemini-3-pro-image} (Vertex AI)            & \na & \texttt{GA}, 28 May 2026 \\
\midrule
\multicolumn{6}{l}{\cellcolor{gray!12}\textit{Open image-editing}} \\
Flux 2                    & \y & \n & \hf{diffusers/FLUX.2-dev-bnb-4bit}                & NF4 4-bit\textsuperscript{b} & \texttt{8775691} (25 Nov 2025) \\
Qwen Image Edit 2511      & \y & \n & \hf{Qwen/Qwen-Image-Edit-2511}                    & bf16 & \texttt{6f3ccc0} \\
Step1X-Edit               & \y & \n & \hf{stepfun-ai/Step1X-Edit-v1p2}                  & bf16 & \texttt{v1p2}\textsuperscript{c} \\
Stable Diffusion 3 Medium & \y & \n & \hf{stabilityai/stable-diffusion-3-medium-diffusers} & bf16 & \texttt{main} \\
FireRed Image Edit        & \y & \n & \makecell[l]{\hf{FireRedTeam/FireRed-Image-Edit-1.0}\\{\footnotesize+\, \hf{FireRed-Image-Edit-1.0-Lightning}} (LoRA)} & bf16 & \makecell[l]{\texttt{1.0}\\{\footnotesize \texttt{Lightning 8-step v1.0}}} \\
\midrule
\multicolumn{6}{l}{\cellcolor{gray!12}\textit{Universal restoration/inpainting; mask models use the EfficientNet-B0 proxy mask}} \\
AutoDIR               & \na & \n & \gh{jiangyitong/AutoDIR}                     & fp32 & \texttt{autodir.ckpt} \\
LanPaint (Qwen/Flux)  & \y  & \y & \gh{scraed/LanPaint}      & bf16 & \texttt{1.5.5}\textsuperscript{d} \\
HYPIR                 & \y  & \y & \makecell[l]{\gh{XPixelGroup/HYPIR}\\{\footnotesize + \hf{lxq007/HYPIR}} (weights)} & fp16 & \makecell[l]{\texttt{HYPIR\_sd2.pth}\\\texttt{base SD-2.1-base}} \\
BIRD                  & \na & \y & \gh{hamadichihaoui/BIRD}                     & fp32 & \texttt{713dab7} \\
\midrule
\multicolumn{6}{l}{\cellcolor{gray!12}\textit{Degradation mask predictor}} \\
EfficientNet-B0 UNet & \na & \na & \textit{trained in this work} & bf16 & \texttt{unet\_efb0\_730k.ckpt} \\
\bottomrule
\end{tabular}
 
\vspace{3pt}
\parbox{0.95\linewidth}{\footnotesize
\textsuperscript{b}\,\texttt{bitsandbytes}\,0.49.2; DiT and text encoder quantised, VAE in bf16; Mistral-3 text encoder.\;
\textsuperscript{c}\,RegionE accelerator disabled; thinking and reflection modes enabled.\;
\textsuperscript{d}\,Qwen and Flux backbones as listed above.\;
\textsuperscript{e}\,LoRA rank 256, $t_{\mathrm{model}}\!=\!t_{\mathrm{coeff}}\!=\!200$.}
\end{table*}
 
We evaluate a diverse set of state of the art closed and open sourced large image editing models' ability in our texture-degraded image restoration task. These include GPT Image 1.5 \cite{openai2025introducing} as of 16 December 2025, Nano Banana (NB) Pro \cite{sun2026can,team2023gemini} as of 28 May 2026, Flux 2 \cite{labs2025flux}, Step1X-Edit \cite{liu2025step1x}, Stable Diffusion 3 \cite{rombach2021highresolution}, Fire Red Image Edit \cite{team2026firered}, Qwen Image Edit \cite{wu2025qwen}. 

We compare the strongest configurations available under our computational and API constraints, chosen according to their state-of-the-art performance in preserving image fidelity and perceptual quality. 
These models leverage their generalizability to various tasks from their large training dataset and prompts to restore images.

Additionally, we compare SOTA models specializing in image restoration and exact inpainting/inverse methods as shown in  \cref{tab:model_config}. These include LanPaint\cite{zheng2025lanpaint} to showcase the versatility of the learned priors of image editing large models, even compared to UIR models. Other models considered are diffusion-based, such as HYPIR \cite{lin2025harnessing}, AutoDIR \cite{jiang2024autodir} and BIRD \cite{chihaoui2024blind}. These either use prompts to identify and restore images based on the degradation type or use partial information from the non-masked regions to infer what the damaged regions should consist of.

\subsection{Prompt Design}

Effective prompt formulation is critical for guiding image editing models to perform faithful mask-localized restoration. Unlike prior image editing restoration benchmarks, which design prompts explicitly instructing models to sharpen, denoise, correct colors, or enhance brightness \cite{sun2026can}, our prompts are deliberately domain-, enhancement-, and degradation-agnostic. We impose no prior assumptions on the color distribution, dynamic range, or any degradation-specific postprocessing instruction for the input image. To study the effect of prompt formulation on restoration quality, we design two prompt variants, as shown in \cref{tab:prompt_comparison}.

We start with a generic, minimal prompt, relying entirely on the model's pretrained priors to solve the task. It imposes no assumptions about what the original content should look like, thus it remains a broadly applicable baseline instruction across scenes, media types, and degradation severities. 
The second is a complex prompt that explicitly frames the task as mask removal and counterfactual content reconstruction. It primes the model on the task and focuses the model's behavior on masked and non-masked regions and what priors it should rely on for filling in masked regions.

Our design instead isolates the most fundamental axis of variation for our task: whether the prompt provides the model with an explicit conceptual framing of the degradation operator, or leaves that inference entirely to the model's pretrained priors. This enables us to analyze the practical sensitivity of the generic prompt and hence model prior information against degradation-explicit prompt as the operative factor driving restoration performance in the art image restoration task.

\begin{table}[t]
\centering
\footnotesize
\caption{Comparison of Generic and Explicit System Prompts for Transparency Textured Image Restoration.}
\label{tab:prompt_comparison}
\vspace{2mm}
\begin{tabular}{p{0.2\columnwidth}p{0.7\columnwidth}}
\toprule
\textbf{Category} & \textbf{Prompt Text} \\
\midrule
Generic & Reconstruct and repair masked image \\
\midrule
Explicit Degradation-Aware & Act like an expert image restoration and enhancement editor. 
\newline Task:\newline
Remove the mask/degradation overlays and reconstruct original image content in the masked regions, as if the mask/degradation had never been applied. Treat the masked regions as corrupted or missing image areas, not as intentional image content. \\
\bottomrule
\end{tabular}
\end{table}

\subsection{Evaluation Metrics}
The model is configured to output an enhanced image at a fixed $512 \times 512$ resolution, heavily conditioning the generative diffusion process on the explicitly defined structural and degradation parameters. Because the generative network outputs a preset fixed-dimension outpainted image, a spatial misalignment occurs with the varying aspect ratios of the native input dataset. To ensure geometric fidelity, standardize image size and allow for pixel-accurate evaluation metrics downstream, we first Lanczos resize to $512$ on the largest side, and then apply a deterministic center-crop on the smallest side to fit the generated outputs on uniform dimensions.

To evaluate restoration performance comprehensively, we compute the average alpha channel value for all grayscale mask overlays and group it into three levels, low, medium, and high mean opacity to emulate task difficulty levels. We follow this by computing full-reference image quality assessment metrics such as L1 (Mean Absolute Error), PSNR, SSIM \cite{wang2004image} to measure pixel-level fidelity and structural consistency, metrics that are particularly challenging for these large image editing models and measuring reconstruction accuracy, and LPIPS \cite{zhang2018unreasonable} for a perceptual measure. 

\section{Experiments}
\begin{table}[htbp]
\centering
\caption{Performance comparison of UNet backbone variants for mask prediction as an image-to-image translation task. Evaluation metrics include Structural Similarity Index Measure (SSIM), Peak Signal-to-Noise Ratio (PSNR), and total model parameter count against a simplified randomly sampled DTD class-wise sample masking the art painting dataset.}
\label{tab:unet_backbone_comparison}
\footnotesize
\begin{tabular}{lccc}
\hline
\textbf{Backbone variant} & \textbf{SSIM} $\uparrow$ & \textbf{PSNR} $\uparrow$ & \textbf{Parameters} \\
\hline
EfficientNet-B0 & \textbf{0.48} & \textbf{16.20} & 730K \\
Ternaus & 0.46 & 15.24 & 45M \\
Swin & 0.42 & 14.41 & 41M \\
Base & 0.43 & 14.29 & 15M \\
\hline
\end{tabular}
\end{table}
\subsection{Dataset Processing}
Ground Truth ($\mathbf{I}_{clean}$): Clean images are sourced from the WikiArt dataset \cite{huggan_wikiart, saleh2015large} where we randomly sample 1000 unique art painting images with a 1:1 aspect ratio, making the dataset landscape painting dominated. We use the images as is without train/val/test splits since its used for benchmarking.

Mask for Input ($\alpha$): A unique degradation process is used. A texture image is randomly selected from the Describable Textures Dataset (DTD). This texture is used as a transparency mask and overlaid onto the clean art image. The resulting textured/masked image serves as the conditional input. The model's task is to remove this texture overlay and restore the original clean painting.

The sizes of the images are Lanczos resized to 512 x 512 resolution.

\subsection{Implementation Details}

The open source models are inferenced with an A100 80GB GPU, with a Unet image to image model trained with a class-stratified DTD mask sampler dataset to retrieve predicted masks from the masked images for the models that have a mask argument (LanPaint, HYPIR and BIRD). We explore various dataset efficient backbones in \cref{tab:unet_backbone_comparison} and select EfficientNet-B0 as the backbone given its performance and smaller model size. By predicting degradation masks we get a mask proxy to make these models work in universal image restoration (UIR) settings with the models, BIRD and LanPaint. The errors in mask prediction are not taken into account for the quantitative and qualitative restoration results, since the focus of the paper is the impact of our prompts on restoration quality.  

The black-box image editing models such as NB Pro and GPT Image 1.5 models are used to query image restoration queries via vertex AI and the openai python-API. NB API requests work best with a 1:1 aspect ratio and 1024 image resolution (below this, it is more likely to produce artifacts) and encode the images into raw bytes as supported by the Genai SDK. To speed up the requests, we process the masked image batches parallelly and aynchronously to work around their available timeslots (due to restrictive requests per minute) for a successful request. Other strategies such as semaphore capping request limits and exponential backoffs on failed requests from unavailability are employed. GPT does not have such strict requirements for the base64 encoded inputs and has less wait time/timeouts between consecutive requests and thus is more reliable with faster times to generate the images. NB and GPT reject requests that violate copyright or those with perceived nudity which can trigger for some of the artistic images in the dataset.

For the open weights models, Flux 2 is loaded from the pre-quantized 4-bit checkpoint:
\linebreak\texttt{diffusers/FLUX.2-dev-bnb-4bit}~\cite{diffusers_flux2_bnb4bit}, which applies 
\linebreak\texttt{bitsandbytes}~\cite{dettmers2023qlora} NF4 quantization, since the full-precision 
model exceeds our compute with 90GB of VRAM. Additionally, this community version uses the Mistral 3 text encoder since the base encoders is not supported with this version. The other models are bfloat16 quantized and have some cpu offloading of layers to effectively utilize the GPU memory. For better performance and speed than the base model, the FireRed Image edit model is loaded with its LoRA adapter. We use the Step1x Edit base configuration without its recent RegionE add-on due to memory constraints. Since RegionE is used for speedups, we consider the version to produce comparable results. 

For the task specific models, HYPIR's underlying Stable Diffusion baseline was selected from its community version \texttt{Manojb/stable-diffusion-2-1-base} \cite{rombach2021highresolution}. 
The BIRD model 
is selected with the inpainting configuration and logic.

\begin{table*}[t]
\centering
\caption{Quantitative benchmarking of generative models for image restoration across different mask opacitys. Lower is better for L1/LPIPS ($\downarrow$), higher is better for PSNR/SSIM ($\uparrow$). We highlight the \best{best} results in yellow and \secondbest{second-best} in red.}
\label{tab:model_mask_opacity}
\setlength{\tabcolsep}{6pt} 
\footnotesize
\begin{tabular}{l l *{8}{S}}
\toprule
 & \textbf{Prompt}& \multicolumn{4}{c}{\textbf{Overall opacity}} & \multicolumn{4}{c}{\textbf{High opacity}} \\
\cmidrule(lr){3-6} \cmidrule(lr){7-10}
\textbf{Model} & \textbf{Type} 
& {L1$\downarrow$} & {LPIPS$\downarrow$} & {PSNR$\uparrow$} & {SSIM$\uparrow$} 
& {L1$\downarrow$} & {LPIPS$\downarrow$} & {PSNR$\uparrow$} & {SSIM$\uparrow$} \\
\midrule

NB Pro \cite{sun2026can}& Simple
    & \secondbest{0.16} & 0.41 & 14.24 & 0.56
    & 0.26              & 0.64 & 9.63  & 0.28  \\
 & Complex
    & \best{0.13}       & \best{0.33} & 15.25 & \secondbest{0.65}
    & \secondbest{0.20} & \secondbest{0.42} & \secondbest{11.87} & \secondbest{0.55}  \\
\midrule
GPT Image 1.5 \cite{openai2025gptimage15} & Simple
    & 0.20 & 0.55 & 12.45 & 0.28
    & 0.26 & 0.60 & 9.76  & 0.27  \\
    & Complex
    & 0.19 & 0.55 & 12.67 & 0.28
    & 0.25 & 0.61 & 10.18 & 0.26  \\
\midrule
Flux 2 \cite{labs2025flux} & Simple
    & 0.20              & 0.42 & 12.42 & 0.37
    & 0.31              & 0.52 & 8.57  & 0.33  \\
    & Complex
    & \secondbest{0.16} & 0.42 & 13.65 & 0.36
    & 0.21              & 0.49 & 11.01 & 0.33  \\
\midrule
Step1x Edit V1P2 \cite{liu2025step1x} & Simple
    & 0.21 & 0.37              & 12.66 & 0.62
    & 0.32 & 0.50              & 8.64  & 0.47  \\
    & Complex
    & 0.20 & \secondbest{0.35} & 13.20 & 0.63
    & 0.32 & 0.49              & 8.68  & 0.48  \\
\midrule
Stable Diffusion 3 Medium \cite{rombach2021highresolution} & Simple
    & 0.21 & 0.61 & 11.76 & 0.34
    & 0.25 & 0.62 & 10.33 & 0.33  \\
    & Complex
    & 0.26 & 0.65 & 10.74 & 0.34
    & 0.38 & 0.70 & 7.18  & 0.30  \\
\midrule
Fire Red Image Edit \cite{team2026firered} & Simple
    & 0.25 & 0.69 & 10.41 & 0.17
    & 0.32 & 0.71 & 8.31  & 0.17  \\
    & Complex
    & 0.23 & 0.69 & 10.69 & 0.17
    & 0.30 & 0.71 & 8.72  & 0.17  \\
\midrule
Qwen Image Edit 2511 \cite{wu2025qwen} & Simple
    & \best{0.13}       & \secondbest{0.35} & \best{15.71}  & \best{0.66}
    & \secondbest{0.20} & \best{0.40}       & \best{12.23}  & \best{0.57}  \\
    & Complex
    & \best{0.13}       & 0.37              & \secondbest{15.51} & \secondbest{0.65}
    & \best{0.19}       & 0.47              & 11.53              & 0.53  \\

\midrule \midrule[\lightrulewidth] 
\textbf{Model} & \textbf{Prompt} & \multicolumn{4}{c}{\textbf{Medium opacity}} & \multicolumn{4}{c}{\textbf{Low opacity}} \\
\cmidrule(lr){3-6} \cmidrule(lr){7-10}
& 
& {L1$\downarrow$} & {LPIPS$\downarrow$} & {PSNR$\uparrow$} & {SSIM$\uparrow$} 
& {L1$\downarrow$} & {LPIPS$\downarrow$} & {PSNR$\uparrow$} & {SSIM$\uparrow$} \\
\midrule

NB Pro \cite{sun2026can} & Simple
    & 0.16              & \secondbest{0.36} & 13.40 & 0.61
    & \best{0.07}       & \secondbest{0.22} & 19.71              & 0.78  \\
 & Complex
    & 0.14              & \best{0.35}       & 14.04 & \secondbest{0.63}
    & \best{0.07}       & \secondbest{0.22} & \secondbest{19.84} & 0.78  \\
\midrule
GPT Image 1.5 \cite{openai2025gptimage15} & Simple
    & 0.20 & 0.55 & 12.08 & 0.28
    & 0.13 & 0.49 & 15.51 & 0.30  \\
    & Complex
    & 0.19 & 0.55 & 12.36 & 0.28
    & 0.13 & 0.49 & 15.48 & 0.30  \\
\midrule
Flux 2 \cite{labs2025flux} & Simple
    & 0.19 & 0.43 & 12.06 & 0.36
    & 0.11 & 0.32 & 16.63 & 0.43  \\
    & Complex
    & 0.14 & 0.43 & 13.93 & 0.36
    & 0.12 & 0.35 & 16.02 & 0.40  \\
\midrule
Step1x Edit V1P2 \cite{liu2025step1x} & Simple
    & 0.21              & 0.38              & 11.62 & 0.60
    & 0.10              & \secondbest{0.22} & 17.72 & \secondbest{0.79}  \\
    & Complex
    & 0.19              & \secondbest{0.36} & 12.36 & 0.61
    & \secondbest{0.09} & \best{0.19}       & 18.57 & \best{0.81}  \\
\midrule
Stable Diffusion 3 Medium \cite{rombach2021highresolution} & Simple
    & 0.25 & 0.62 & 10.33 & 0.33
    & 0.14 & 0.58 & 14.63 & 0.37  \\
    & Complex
    & 0.25 & 0.64 & 10.35 & 0.33
    & 0.14 & 0.60 & 14.69 & 0.38  \\
\midrule
Fire Red Image Edit \cite{team2026firered} & Simple
    & 0.24 & 0.69 & 10.36 & 0.16
    & 0.18 & 0.66 & 12.55 & 0.17  \\
    & Complex
    & 0.22 & 0.69 & 10.72 & 0.16
    & 0.17 & 0.67 & 12.64 & 0.17  \\
\midrule
Qwen Image Edit 2511 \cite{wu2025qwen} & Simple
    & \secondbest{0.13} & 0.39 & \best{15.13}       & \best{0.64}
    & \best{0.07}       & 0.27 & 19.76              & 0.77  \\
    & Complex
    & \best{0.12}       & 0.39 & \secondbest{15.03} & 0.62
    & \best{0.07}       & 0.24 & \best{19.98}       & \secondbest{0.79}  \\
\midrule

\bottomrule
\label{tab:model_mask_opacity}
\end{tabular}
\end{table*}

\subsection{Quantitative Results}
\label{sec:quant_res}

Table \ref{tab:model_mask_opacity} compares large image editing models for restoring the art dataset images with our degradation task masks across different mask level opacities under both generic and explicit degradation-aware prompt conditions. Under generic prompting, Qwen Image Edit 2511 leads across the majority of metrics, achieving the best L1 (0.13), LPIPS (0.35), PSNR (15.71 dB), and SSIM (0.66) under generic prompting, with Nano Banana (NB) Pro (Complex) closely competitive, attaining the best LPIPS outright (0.33) and matching Qwen's L1. Notably, prompt complexity has a measurable and consistent positive effect on restoration quality at higher opacity levels. Models such as NB Pro and Flux 2 show meaningful L1 and PSNR improvements when switching to explicit prompts (e.g. NB Pro improves by 2.24 dB PSNR), suggesting that richer semantic context in the prompt provides additional signal that partially compensates for the reduced unmasked area available to the model. Examining the prompt dimension more closely, the models do not respond to prompt complexity uniformly. They separate into four distinct response regimes. NB Pro shows positive responses particularly at high opacity with a structural gain of 0.27, Flux 2 is an opacity-dependent responder with a +2.44 dB jump at high opacity, but -0.61 dB at low opacity. Furthermore, SD3 Medium shows negative response with more explicit degradation-aware prompts, while Qwen Image Edit is mostly insensitive to prompt complexity, with slightly better performance with generic prompts.

Under high opacity, Qwen  obtains the best high-opacity result on three of four metrics, LPIPS (0.40), PSNR (12.23 dB), and SSIM (0.57) under generic prompts, while NB2 Pro Complex secures second-best positions across all four metrics at this level. The overall opacities represent the average restoration evaluation metrics across all three opacity levels. As mask opacity decreases to medium and low levels, the performance landscape shifts and inter-model variance narrows, though ranking trends are broadly preserved. At medium opacity, Qwen Image Edit 2511 retains the best PSNR and SSIM (15.13 dB, 0.64 under generic prompts), while NB Pro Complex achieves the best LPIPS (0.35), indicating stronger perceptual quality preservation when more of the image remains unmasked. At low opacity , performance converges across models, with NB Pro, Qwen, and Step1x all achieving L1 of 0.07–0.09. Step1x Edit Complex emerges as the standout at this level, recording the best LPIPS (0.19) and SSIM (0.81), while Qwen Complex achieves the best PSNR (19.98 dB) and NB Pro Complex the second-best (19.84 dB). GPT Image 1.5 and FireRed Image Edit consistently underperform relative to the field across all opacity levels, with FireRed in particular showing near-invariant SSIM values regardless of opacity or prompt type, indicative of poor structural fidelity. Within the present, non-uniform implementation protocol, several image-editing models obtain higher scores than the UIR-specific models such as AutoDIR, HYPIR, and BIRD, which plateau at lower PSNR and SSIM ceilings.

\begin{table*}[t]
\centering
\caption{Quantitative benchmarking of generative models for image restoration across different mask opacitys. Lower is better for L1/LPIPS ($\downarrow$), higher is better for PSNR/SSIM ($\uparrow$). We highlight the \best{best} results in yellow  and \secondbest{second-best} in red.}
\label{tab:IR_model_mask_opacity}
\footnotesize
\setlength{\tabcolsep}{6pt}
\begin{tabular}{l l *{8}{S}}
\toprule
 & \textbf{Prompt}& \multicolumn{4}{c}{\textbf{overall opacity}} & \multicolumn{4}{c}{\textbf{High opacity}} \\
\cmidrule(lr){3-6} \cmidrule(lr){7-10}
\textbf{Model} & \textbf{Prompt Type} 
& {L1$\downarrow$} & {LPIPS$\downarrow$} & {PSNR$\uparrow$} & {SSIM$\uparrow$} 
& {L1$\downarrow$} & {LPIPS$\downarrow$} & {PSNR$\uparrow$} & {SSIM$\uparrow$} \\
\midrule
LanPaint-Qwen \cite{zheng2025lanpaint}    & Simple                  & 0.25 & 0.43 & 11.19 & 0.57 & 0.38 & 0.57 & 7.29  & 0.41  \\
     & Complex                 & \secondbest{0.24} & 0.43 & \secondbest{11.25} & 0.57 & \secondbest{0.37} & 0.57 & \secondbest{7.44}  & 0.41  \\
\midrule
LanPaint-Flux \cite{zheng2025lanpaint}    & Simple                  & 0.25 & \secondbest{0.42} & 11.18 & \best{0.58} & 0.38 & \secondbest{0.56} & 7.25  & \secondbest{0.42}  \\
     & Complex                 & 0.25 & \secondbest{0.42} & 11.18 & \best{0.58} & 0.38 & \secondbest{0.56} & 7.25  & \secondbest{0.42}  \\
\midrule
BIRD \cite{chihaoui2024blind}          & N/A                  & 0.3 & 0.62 & 9.12 & 0.35 & 0.41 & 0.70  & 6.66  & 0.33  \\
\midrule
HYPIR \cite{lin2025harnessing}          & Simple                  & 0.25  & 0.57 & 10.82 & 0.36 & 0.38 & 0.66 & 7.17 & 0.29  \\
          & Complex                 & 0.25 & 0.56 & 10.83 & 0.36 & 0.38 & 0.65 & 7.17 & 0.29  \\
\midrule
AutoDIR \cite{jiang2024autodir}          & N/A                  & \best{0.23}  & \best{0.41} & \best{11.74} & \best{0.58} & \best{0.34} & \best{0.54}  & \best{7.89}  & \best{0.43}  \\
\midrule \midrule[\lightrulewidth]
\textbf{Model} & \textbf{Prompt} & \multicolumn{4}{c}{\textbf{Medium opacity}} & \multicolumn{4}{c}{\textbf{Low opacity}} \\
\cmidrule(lr){3-6} \cmidrule(lr){7-10}
& 
& {L1$\downarrow$} & {LPIPS$\downarrow$} & {PSNR$\uparrow$} & {SSIM$\uparrow$} 
& {L1$\downarrow$} & {LPIPS$\downarrow$} & {PSNR$\uparrow$} & {SSIM$\uparrow$} \\
\midrule
LanPaint-Qwen \cite{zheng2025lanpaint}    & Simple                  & \secondbest{0.24} & \best{0.42} & 10.58 & 0.56 & \secondbest{0.12} & \secondbest{0.29} & 15.70  & 0.73 \\
     & Complex                 & \secondbest{0.24} & \best{0.42} & \secondbest{10.60}  & 0.56 & \secondbest{0.12} & \secondbest{0.29} & 15.70  & 0.73 \\
\midrule
LanPaint-Flux \cite{zheng2025lanpaint}    & Simple                  & \secondbest{0.24} & \best{0.42} & 10.58 & \best{0.57} & \secondbest{0.12} & \secondbest{0.29} & \secondbest{15.72} & \secondbest{0.74} \\
     & Complex                 & \secondbest{0.24} & \best{0.42} & 10.58 & \best{0.57} & \secondbest{0.12} & \secondbest{0.29} & \secondbest{15.72} & \secondbest{0.74} \\
\midrule
BIRD \cite{chihaoui2024blind}          & N/A                  & 0.25 & 0.54 & 10.52 & 0.46 & 0.25 & 0.61 & 10.18 & 0.26  \\
\midrule
HYPIR \cite{lin2025harnessing}          & Simple                  & 0.25  & 0.56 & 10.34 & 0.36 & 0.13 & 0.49  & 14.96 & 0.44  \\
          & Complex                 & 0.25 & 0.56 & 10.34 & 0.36 & 0.13 & 0.48 & 14.97 & 0.44  \\
\midrule
AutoDIR \cite{jiang2024autodir}          & N/A                  & \best{0.23}  & \best{0.42} & \best{10.96} & \best{0.57} & \best{0.11} & \best{0.27}  & \best{16.38}  & \best{0.75}  \\
\midrule
\bottomrule
\end{tabular}
\end{table*}

\cref{tab:IR_model_mask_opacity} benchmarks four universal image restoration (UIR) models under the same opacity levels and prompt types. The vanilla LanPaint implementation models prioritize the input masks over the text prompts for the image restorations. The LanPaint variants (Qwen and Flux) are competitive in LPIPS and SSIM, where LanPaint-Flux achieves tied-best SSIM at overall opacity (0.58) and LanPaint-Qwen attains the best LPIPS at medium opacity (0.42). These minimal gains for prompt types are consistent for the models listed in the table. To emphasize the importance of image factors in these models, we see that AutoDIR is strongest overall among restoration-specific models, recording the best L1, PSNR, and SSIM at overall opacity (0.23, 11.74 dB, 0.58) and sweeping all four metrics at low opacity (L1: 0.11, LPIPS: 0.27, PSNR: 16.38 dB, SSIM: 0.75). The models relying more on image priors, such as BIRD and HYPIR, underperform across all opacities, since they require binary masks and known degradations, respectively.

\subsection{Qualitative Results}
\label{sec:qual_res}
\begin{figure*}[t]
    \centering

    \begin{tabular}{cccccc}

        \includegraphics[width=0.13\textwidth]{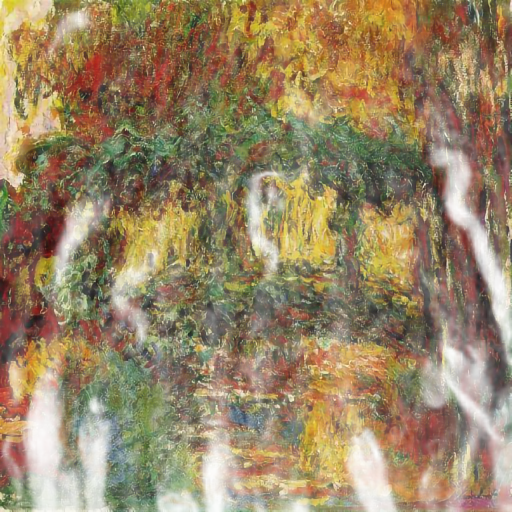} &
        \includegraphics[width=0.13\textwidth]{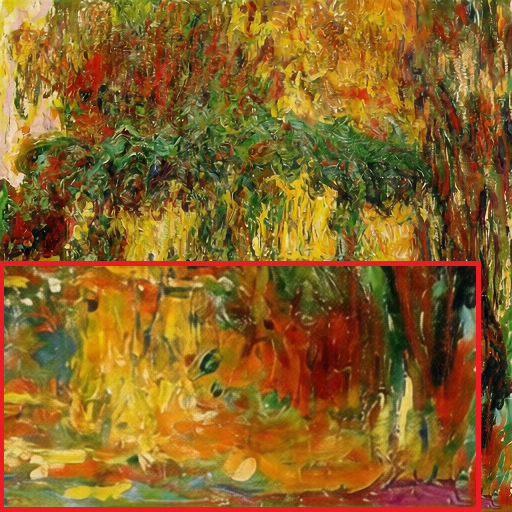} &
        \includegraphics[width=0.13\textwidth]{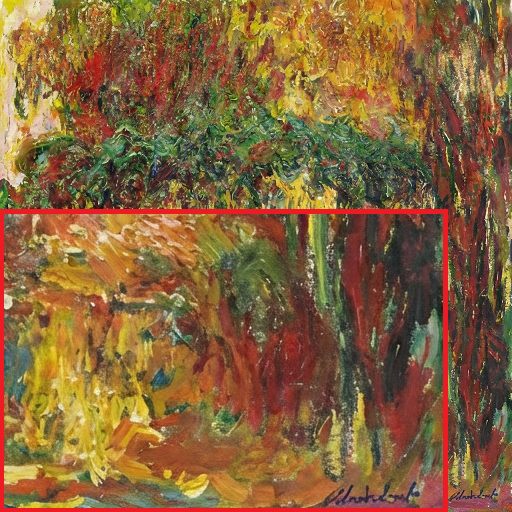} &
        \includegraphics[width=0.13\textwidth]{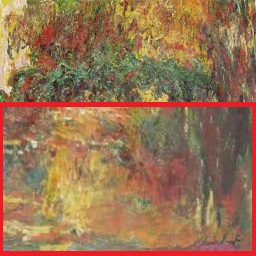} &
        \includegraphics[width=0.13\textwidth]{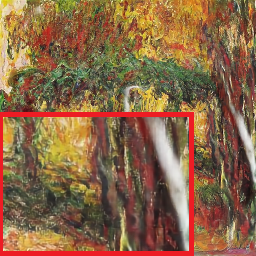} &
        \includegraphics[width=0.13\textwidth]{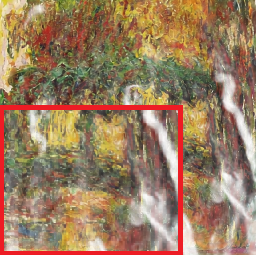} 
        \\ \vspace{0.5ex}
        Low opacity & Qwen Image Edit & NB Pro & Step1x Edit & AutoDIR & LanPaint(Qwen) \\

        \includegraphics[width=0.13\textwidth]{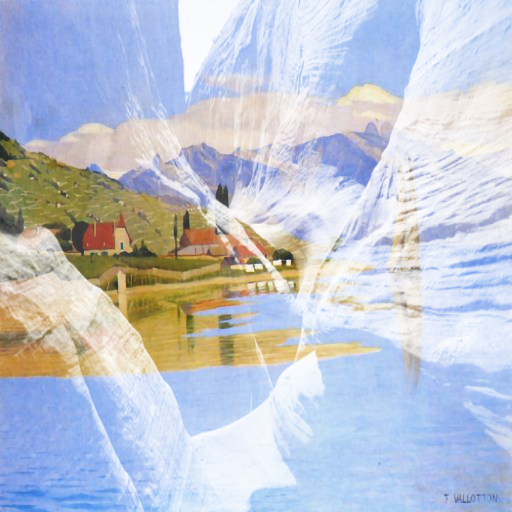} &
        \includegraphics[width=0.13\textwidth]{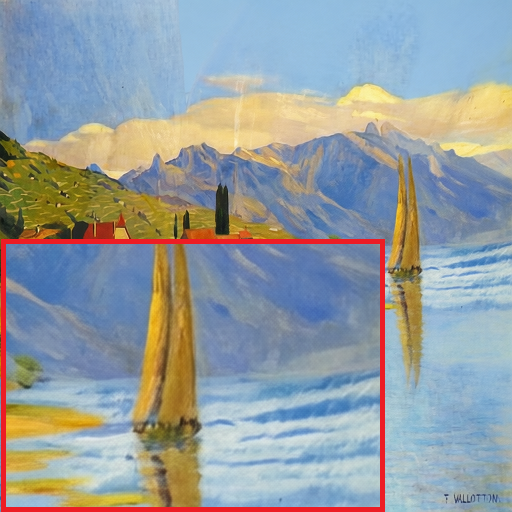} &
        \includegraphics[width=0.13\textwidth]{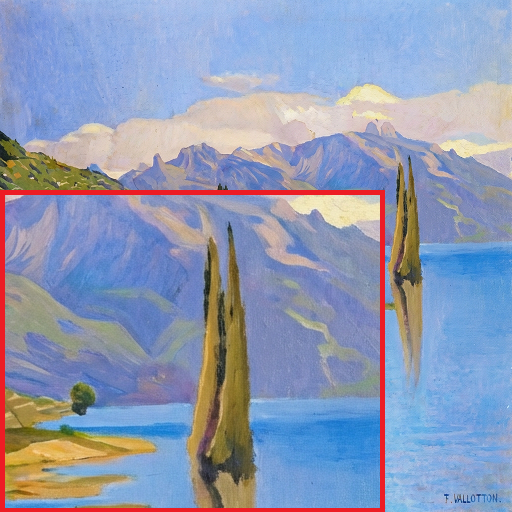} &
        \includegraphics[width=0.13\textwidth]{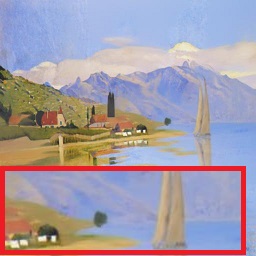} &
        \includegraphics[width=0.13\textwidth]{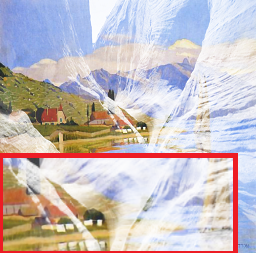} &
        \includegraphics[width=0.13\textwidth]{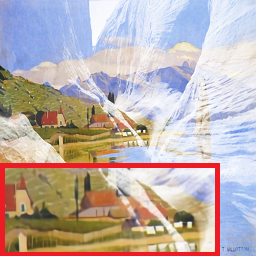} \\ \vspace{0.5ex}
        Medium opacity & Qwen Image Edit & NB Pro & Step1x Edit & AutoDIR & LanPaint(Qwen) \\

        \includegraphics[width=0.13\textwidth]{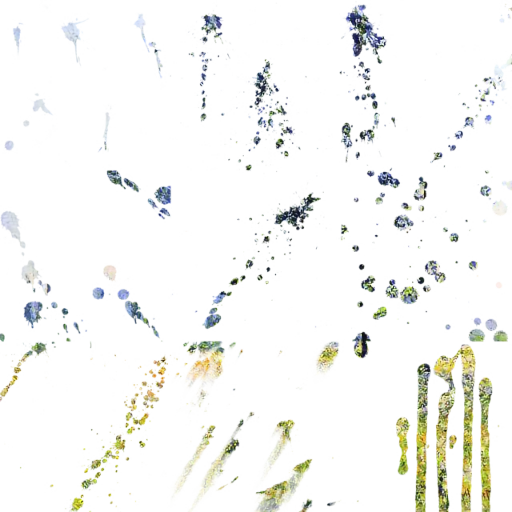} &
        \includegraphics[width=0.13\textwidth]{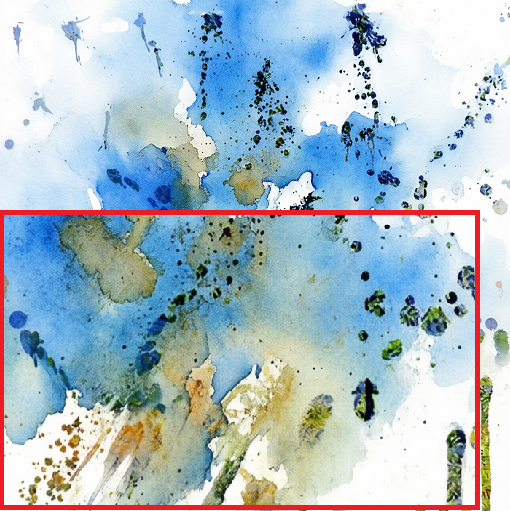} &
        \includegraphics[width=0.13\textwidth]{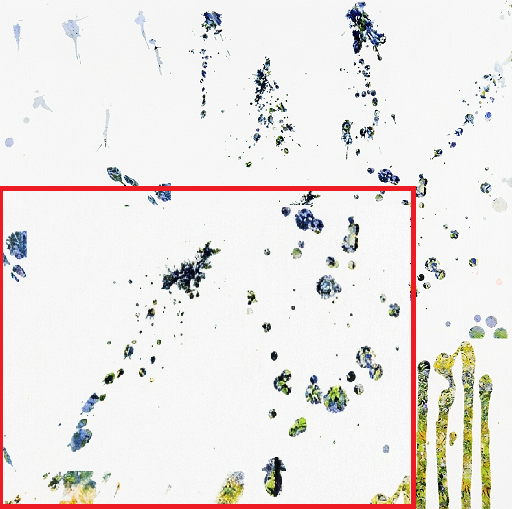} &
        \includegraphics[width=0.13\textwidth]{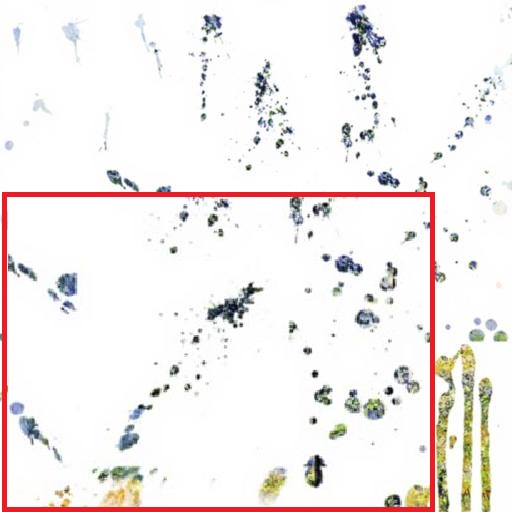} &
        \includegraphics[width=0.13\textwidth]{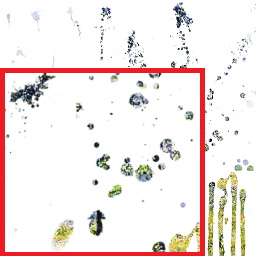} &
        \includegraphics[width=0.13\textwidth]{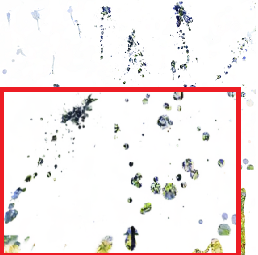}  \\ \vspace{0.5ex}
        High opacity & Qwen Image Edit & NB Pro & Step1x Edit & AutoDIR & LanPaint(Qwen) \\

    \end{tabular}

    \caption{Visual comparison of image restoration results on three sample images across different alpha texture mask opacitys. The Masked image is shown in the first column, followed by predictions from the models, Qwen Image Edit, Nano Banana (NB) Pro, Step1x Edit, AutoDIR, and LanPaint with Qwen backbone.}
    \label{fig:alpha_level_comparison_grid}
\end{figure*}

Fig. \ref{fig:alpha_level_comparison_grid} presents a visual comparison of the best models in the quantitative experiments section under image editing models and image restoration-specific models under the three alpha texture mask opacity levels, zooming in on regions with the most inconsistencies with the ground truth images. At low opacity, most models produce visually coherent reconstruction. Qwen Image Edit recovers plausible color and texture but introduces tonal shifts inconsistent with the ground truth palette, NB adds additional details not present in the ground truth image. AutoDIR maintains the highest structural fidelity to the degraded input under low-opacity but produces residual artifacts rather than semantically coherent completions. LanPaint (Qwen) similarly preserves global layout but exhibits incomplete inpainting in localized high-frequency regions, highlighted by the red bounding boxes, which draw attention to regions of notable reconstruction failure.

As opacity increases to medium and high levels, the performance gap between generative and fidelity-preserving approaches widens substantially. Under medium opacity, Step1x Edit and Qwen Image Edit maintain scene-level plausibility, recovering buildings, sky, water, and vegetation structure but diverge from ground truth in fine structural details such as sailboat structure and mountain silhouette, reflecting the tendency of image editing models to substitute semantically probable completions for accurate reconstruction without using the translucent underlying information. NB Pro degrades more sharply with opacity, producing color bleeds and structural incoherence particularly in regions with complex textures. Under high opacity, all models face severe reconstruction difficulty, as the remaining unmasked signal is insufficient to anchor scene-level semantics darkening and slightly diluting the splatter regions. Qwen Image Edit is the only model that tries to extrapolate possible reconstructions with the large missing region.


\newcolumntype{C}[1]{>{\centering\arraybackslash}m{#1}}

\begin{figure*}[t]
    \centering
    \setlength{\tabcolsep}{2pt}

    \begin{tabular}{@{}C{1.5em}*{6}{C{0.14\textwidth}}@{}}
        &
        Masked Image &
        Qwen Image Edit &
        NB Pro &
        Flux 2 &
        SD3 Medium &
        Ground Truth
        \\

        \rotatebox[origin=c]{90}{\strut Simple} &
        \includegraphics[width=\linewidth]
        {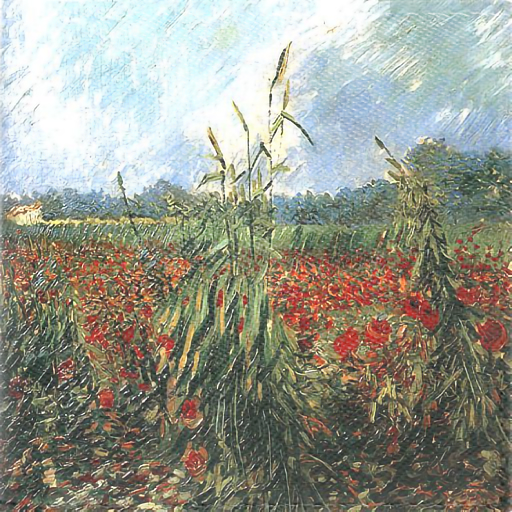} &
        \includegraphics[width=\linewidth]
        {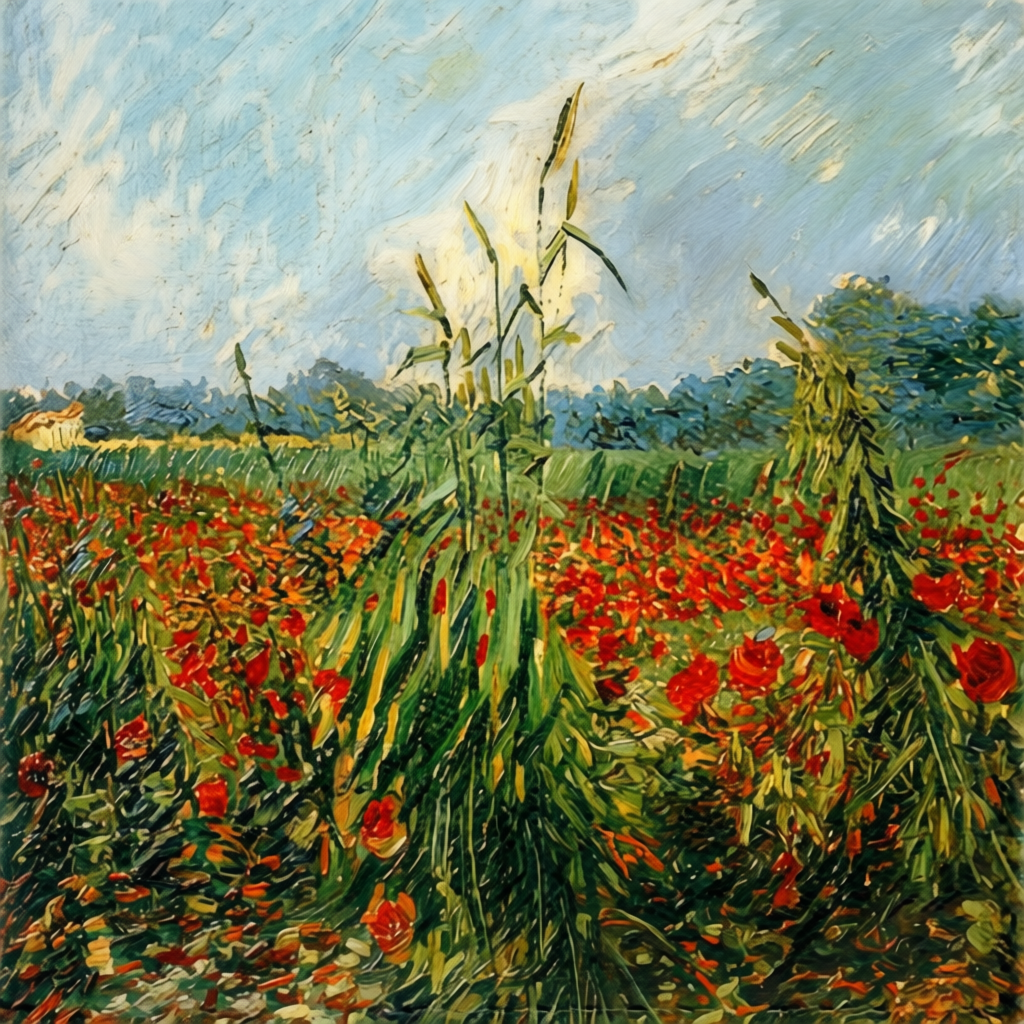} &
        \includegraphics[width=\linewidth,height=0.14\textwidth]
        {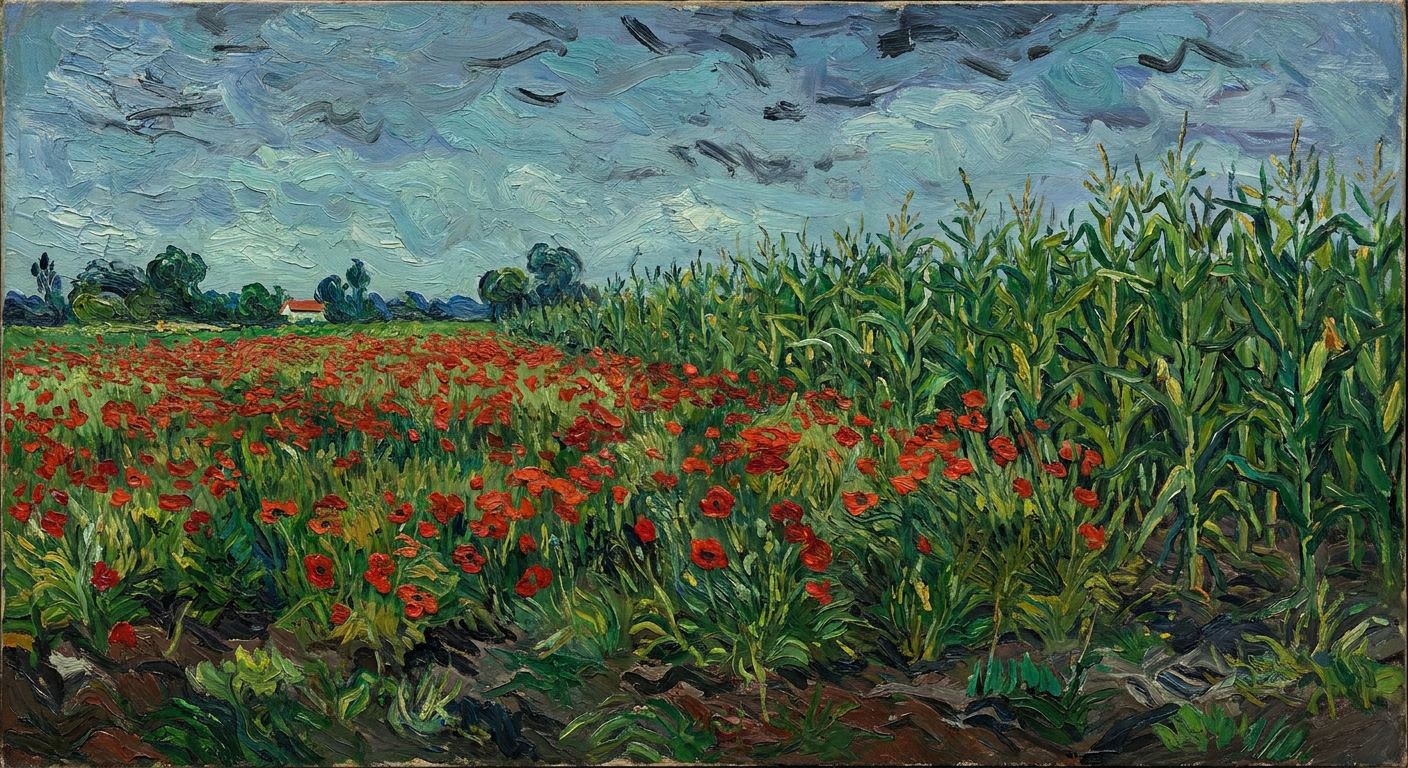} &
        \includegraphics[width=\linewidth]
        {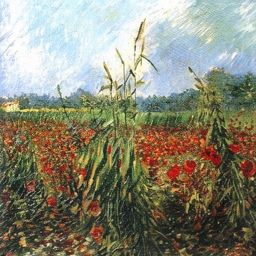} &
        \includegraphics[width=\linewidth]
        {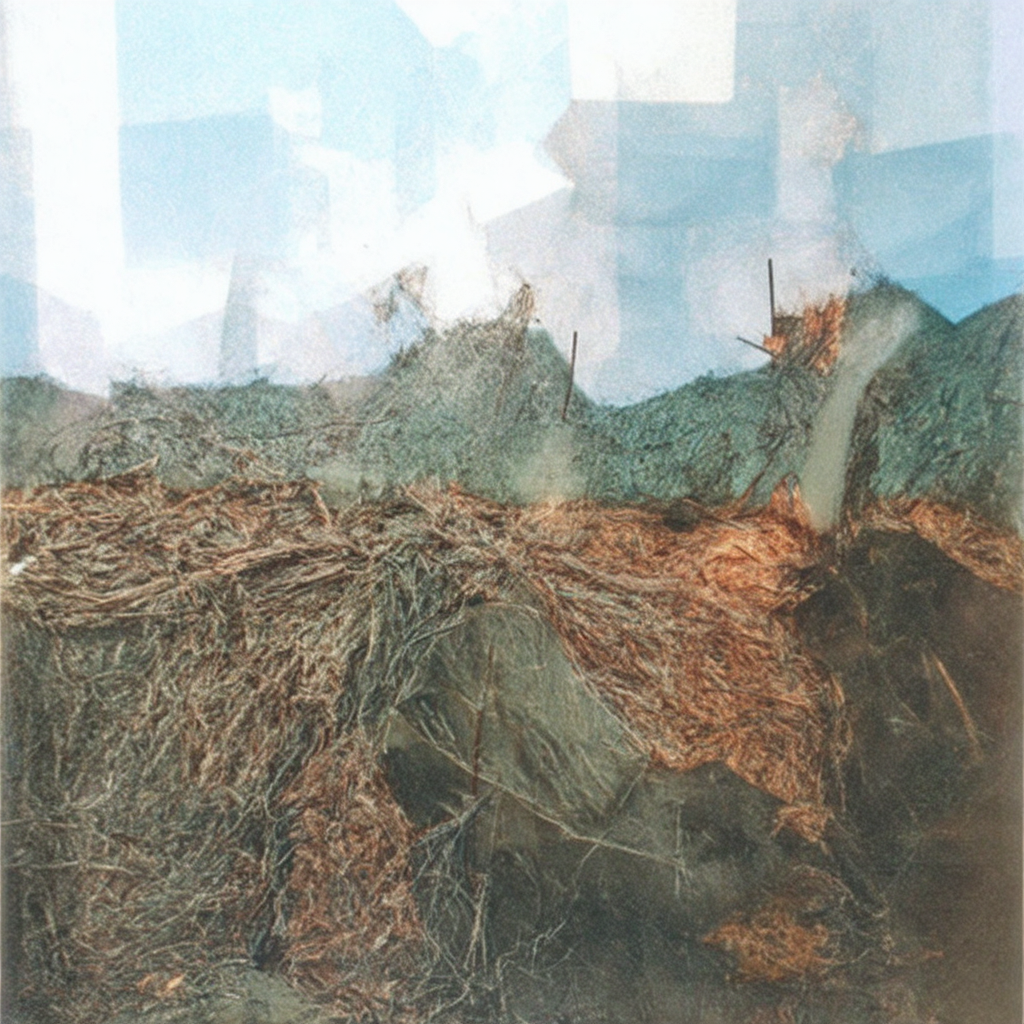} &
        \includegraphics[width=\linewidth]
        {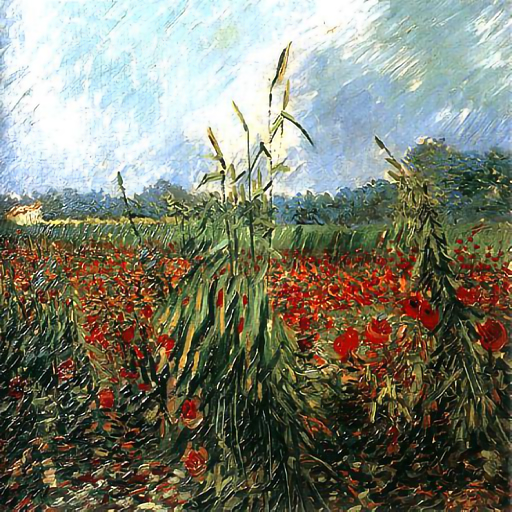}
        \\[0.5ex]

        \rotatebox[origin=c]{90}{\strut Complex} &
        \includegraphics[width=\linewidth]
        {figures/qual_prompt_type/masked_13caabb93203ddcd0b7bb4c918b0ff78.png} &
        \includegraphics[width=\linewidth]
        {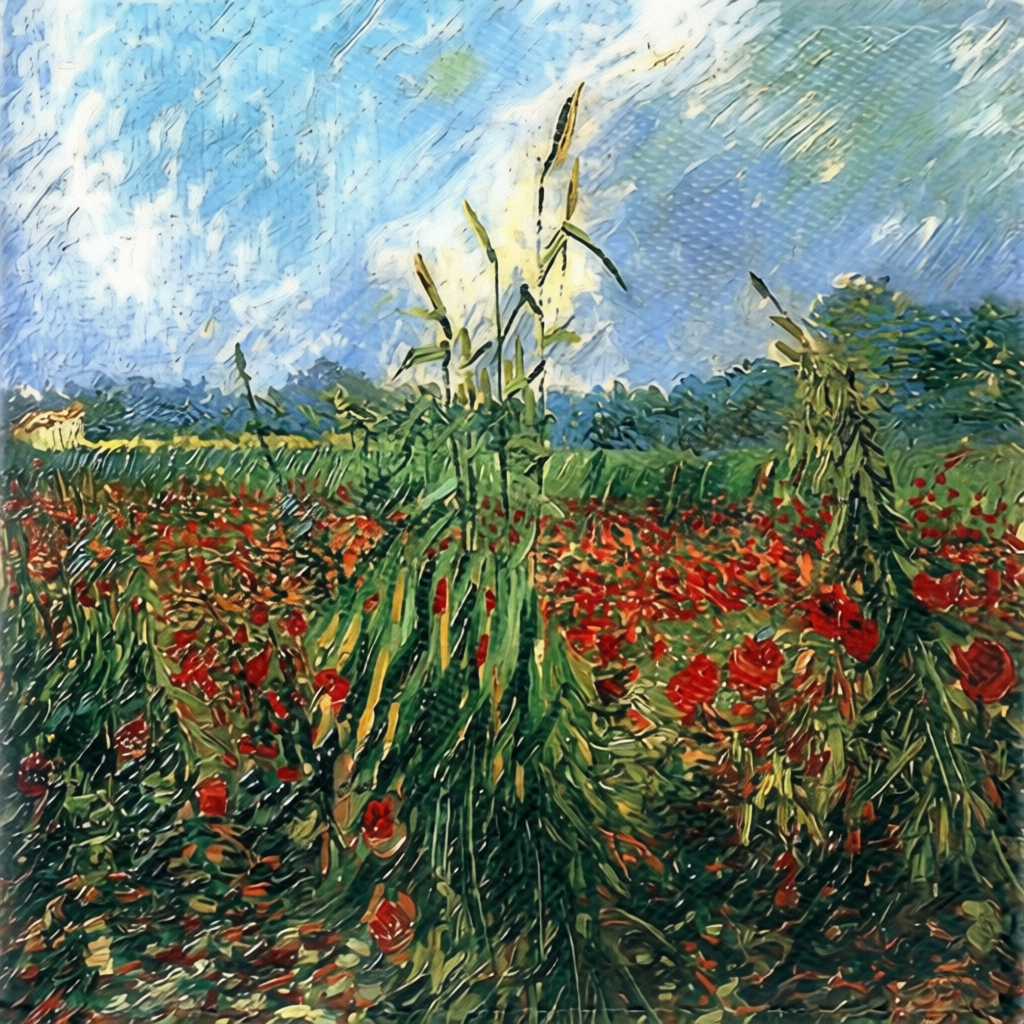} &
        \includegraphics[width=\linewidth]
        {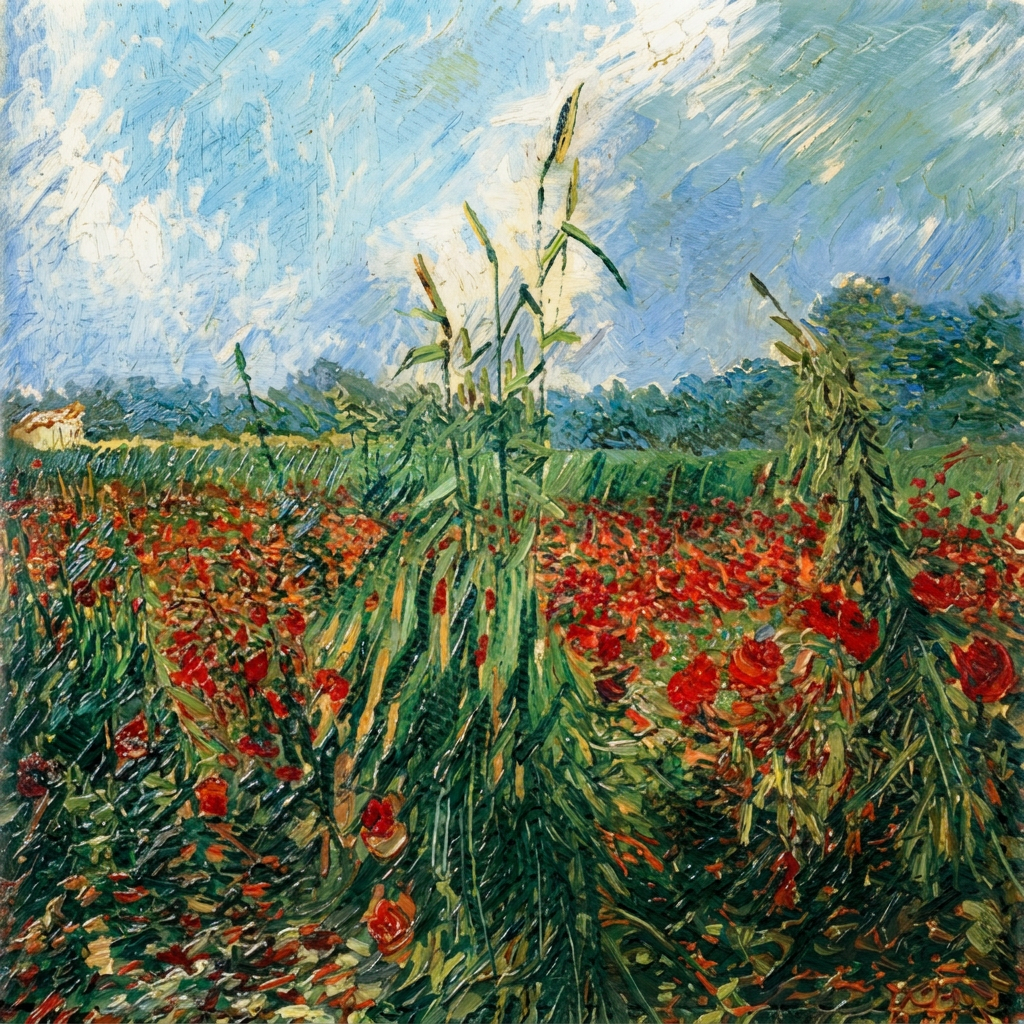} &
        \includegraphics[width=\linewidth]
        {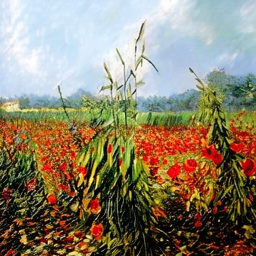} &
        \includegraphics[width=\linewidth]
        {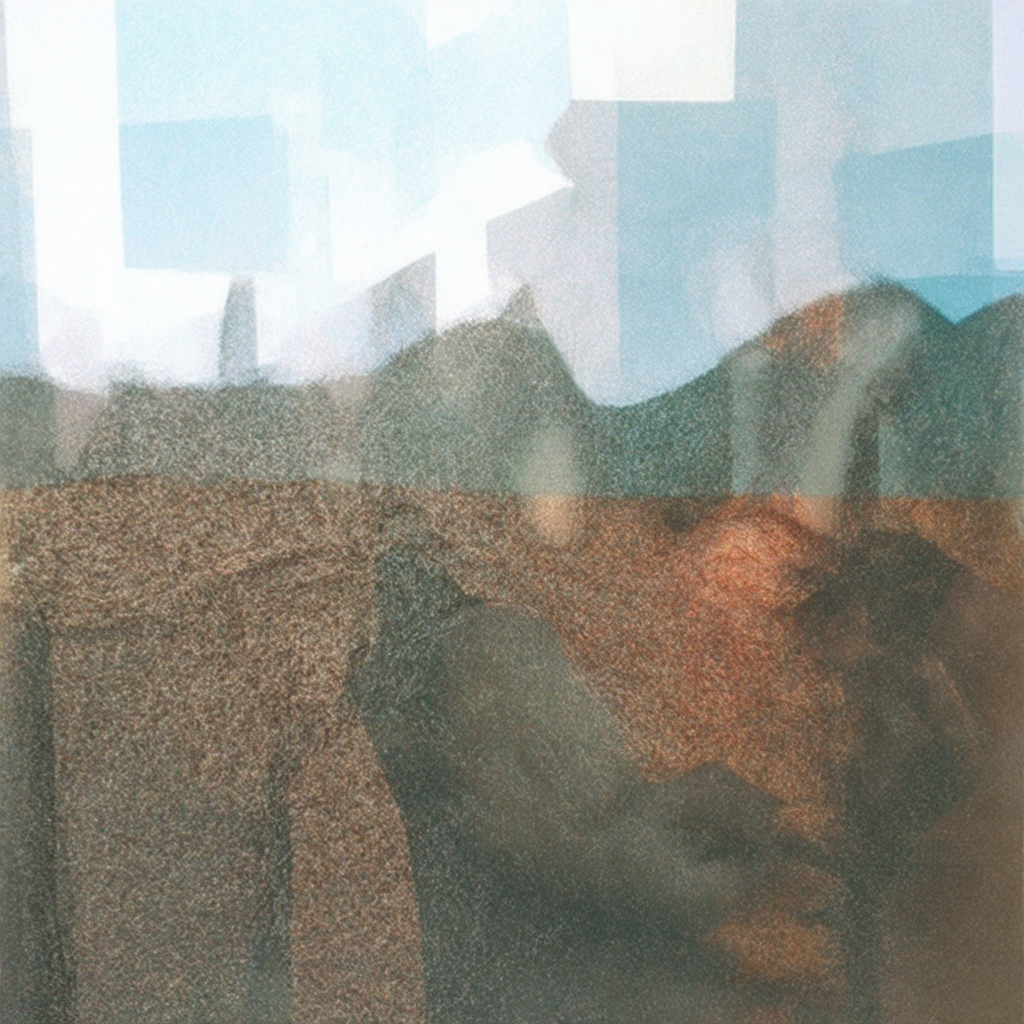} &
        \includegraphics[width=\linewidth]
        {figures/qual_prompt_type/13caabb93203ddcd0b7bb4c918b0ff78.png}
    \end{tabular}
\caption{Visual comparison of image restoration results on two sample images across different prompt types for image editing models. The first column explores restorations under generic prompts and the second explores explicit degradation-aware prompts. The Masked image is shown in the first column, followed by predictions from the models, NB Pro, Flux 2, SD3 Medium, and Qwen Image Edit .}
    \label{fig:prompt_type_comparison_grid}
\end{figure*}

Fig. \ref{fig:prompt_type_comparison_grid} explores the variations of restoration quality with different prompts for a low-opacity sample, where the model cannot rely on local texture cues to distinguish degradation from the intended content and needs to focus on the task instruction to resolve this ambiguity. The models are selected from \cref{tab:model_mask_opacity} based on prompt sensitivity. Most models, aside from NB, have more deviation from the ground truth with the explicit degradation-aware prompt, with the instructions producing lower deviations in the model output. There is less universal transferability of prompts across these image editing models for image restoration. 

Since the explicit prompt explicitly frames the masked regions as corrupted content, the semantic content of the images are retained for NB, showing the model and opacity dependent effects based on task framing, especially at high opacity. For the other models, the explicit prompts seems to make the outputs more reliant on the model priors as compared to the input images, showing more separation of the mask from the content. There is notably more effect on the mask opacity and type as compared to the prompts, but from the above mentioned table, the prompts has conditional effect on the restoration quality depending on the mask opacity. To elaborate, they synthesize corrupt regions in excess and hurt the reconstruction quality for lower opacity masks, especially in the case of SD3 Medium. Flux 2 and Qwen Image Edit increase the contrast of the resulting output, while SD3 attenuates and supersaturates the result.

\begin{figure*}[!htp]
    \centering

    \begin{tabular}{cccccc}
        Damaged Image & NB Pro & AutoDIR & Step1x Edit & LanPaint(Qwen) & Qwen Image Edit \\

        \includegraphics[width=0.13\textwidth]{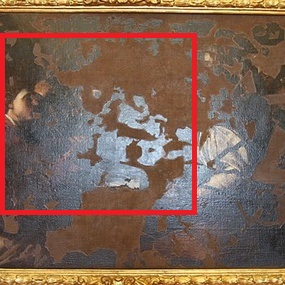} &
        \includegraphics[width=0.13\textwidth]{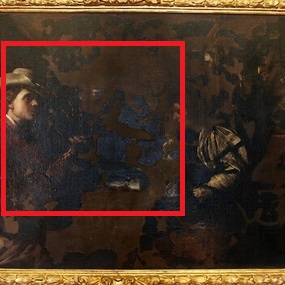} &
        \includegraphics[width=0.13\textwidth,height = 0.13\textwidth]{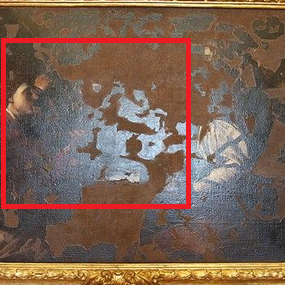} &
        \includegraphics[width=0.13\textwidth]{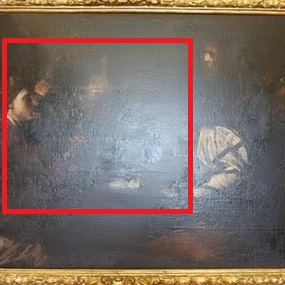} &
        \includegraphics[width=0.13\textwidth]{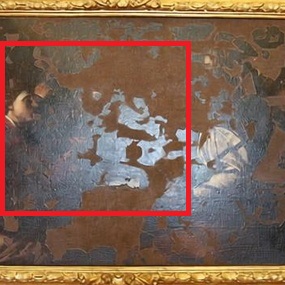} & 
        \includegraphics[width=0.13\textwidth]{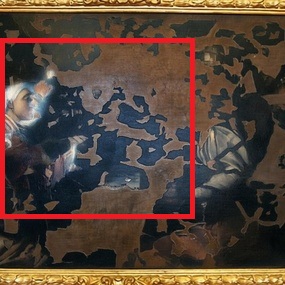} \\ \vspace{0.5ex}
        
        \includegraphics[width=0.13\textwidth]{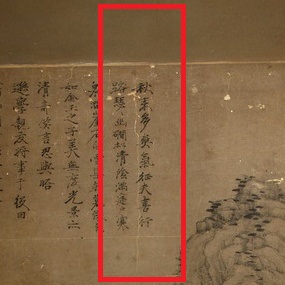} &
        \includegraphics[width=0.13\textwidth]{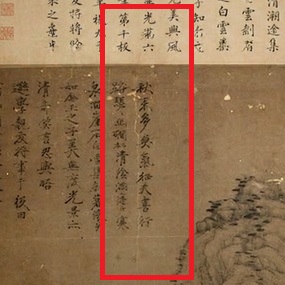} &
        \includegraphics[width=0.13\textwidth,height = 0.13\textwidth]{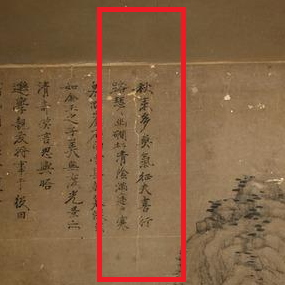} &
        \includegraphics[width=0.13\textwidth]{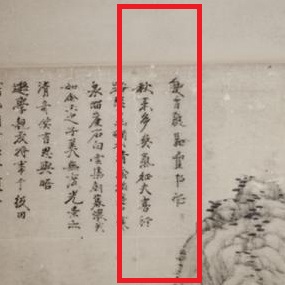} &
        \includegraphics[width=0.13\textwidth]{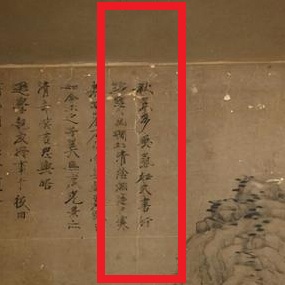} & 
        \includegraphics[width=0.13\textwidth]{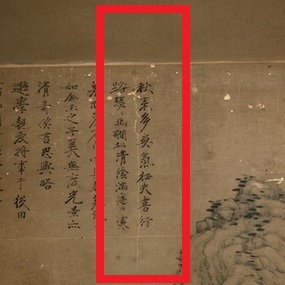} \\ \vspace{0.5ex}

        \includegraphics[width=0.13\textwidth]{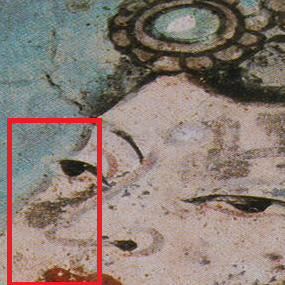} &
        \includegraphics[width=0.13\textwidth]{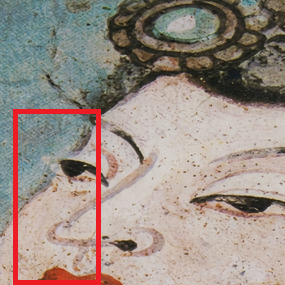} &
        \includegraphics[width=0.13\textwidth]{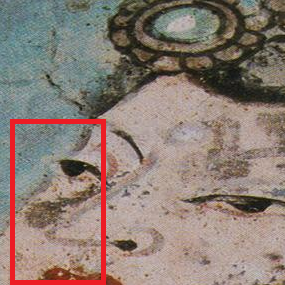} &
        \includegraphics[width=0.13\textwidth]{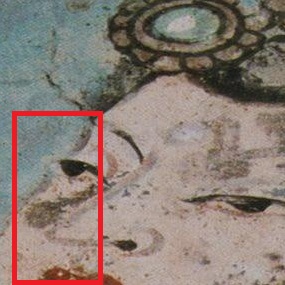} &
        \includegraphics[width=0.13\textwidth]{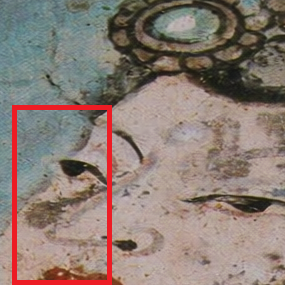} & 
        \includegraphics[width=0.13\textwidth]{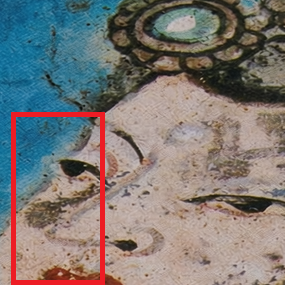} \\ \vspace{0.5ex}

    \end{tabular}

    \caption{Visual comparison of image restoration results on sample images from multiple datasets \cite{artworks_damaged_undamaged, heritage_paintings_restoration, xu2024comprehensive} with real world degradations. The Masked image is shown in the first column, followed by predictions from the models, NB Pro, AutoDIR, Step1x Edit, LanPaint, and Qwen Image Edit .}
    \label{fig:realworld_samples}
\end{figure*}

\begin{figure}
    \centering

    \begin{tabular}{ccc}
        \textbf{Degraded Image} & \textbf{Restored image} & \textbf{Ground Truth} \\

        \includegraphics[width=0.12\textwidth]{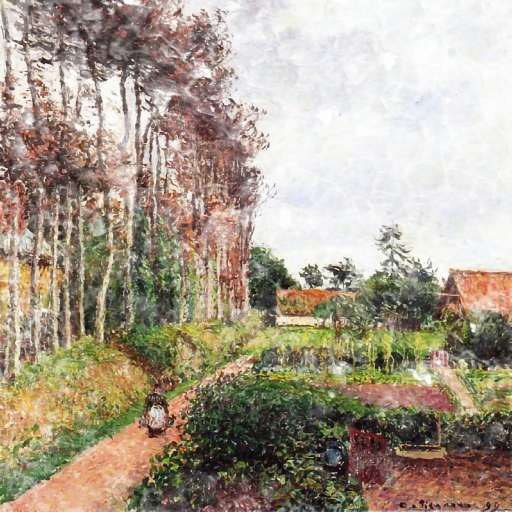} &
        \includegraphics[width=0.12\textwidth]{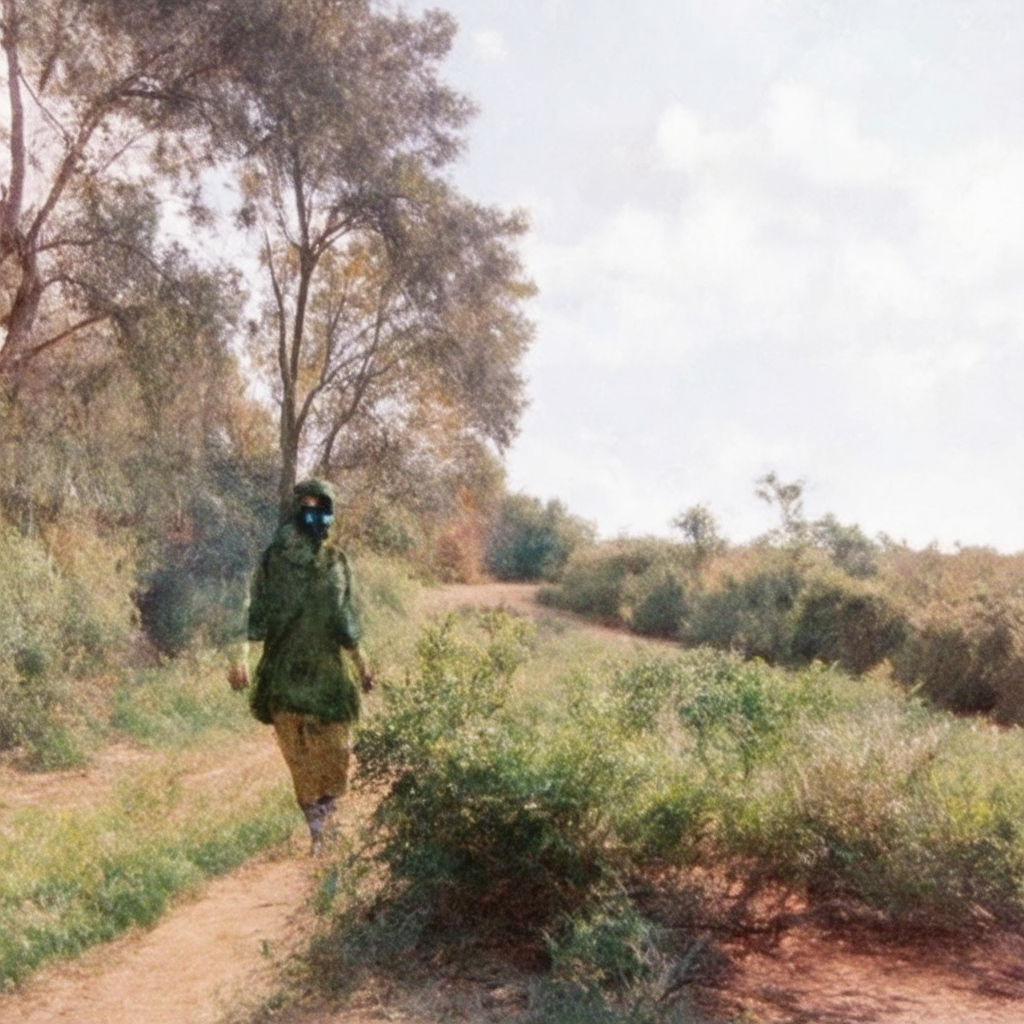} &
        \includegraphics[width=0.12\textwidth]{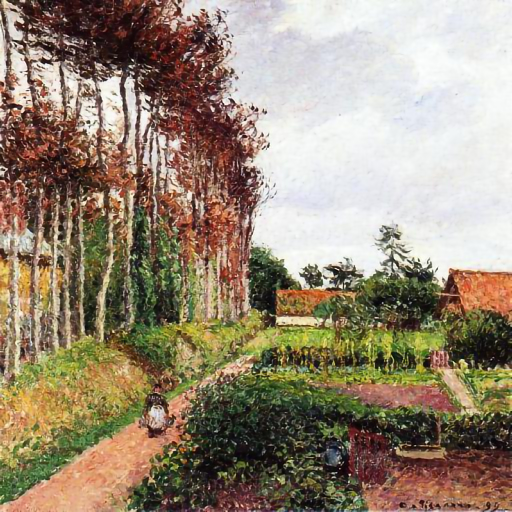} \\
        Low opacity & SD3 Medium & \\
        \includegraphics[width=0.12\textwidth]{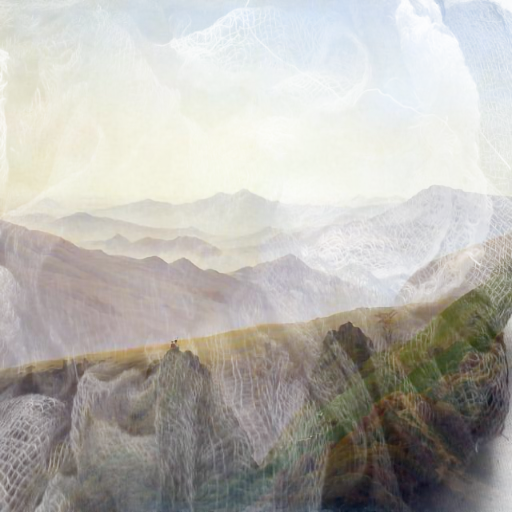} &
        \includegraphics[width=0.12\textwidth]{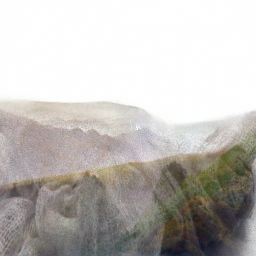} &
        \includegraphics[width=0.12\textwidth]{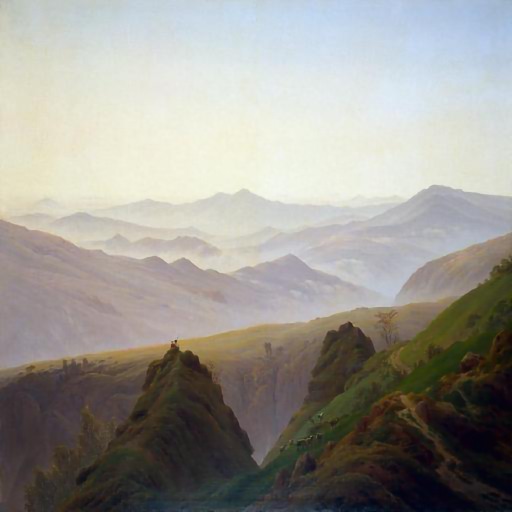} \\
        Low opacity & BIRD & \\
        \includegraphics[width=0.12\textwidth]{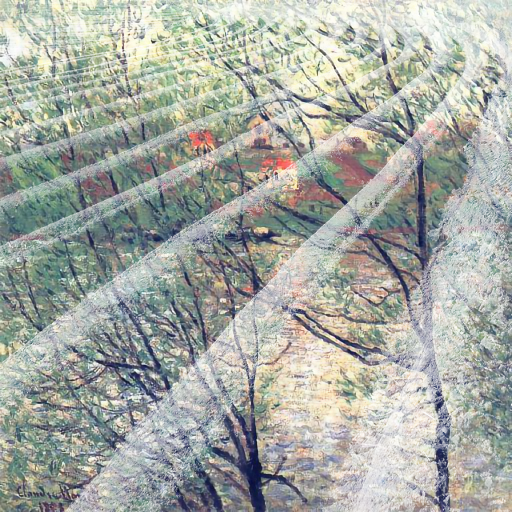} &
        \includegraphics[width=0.12\textwidth,height = 0.12\textwidth]{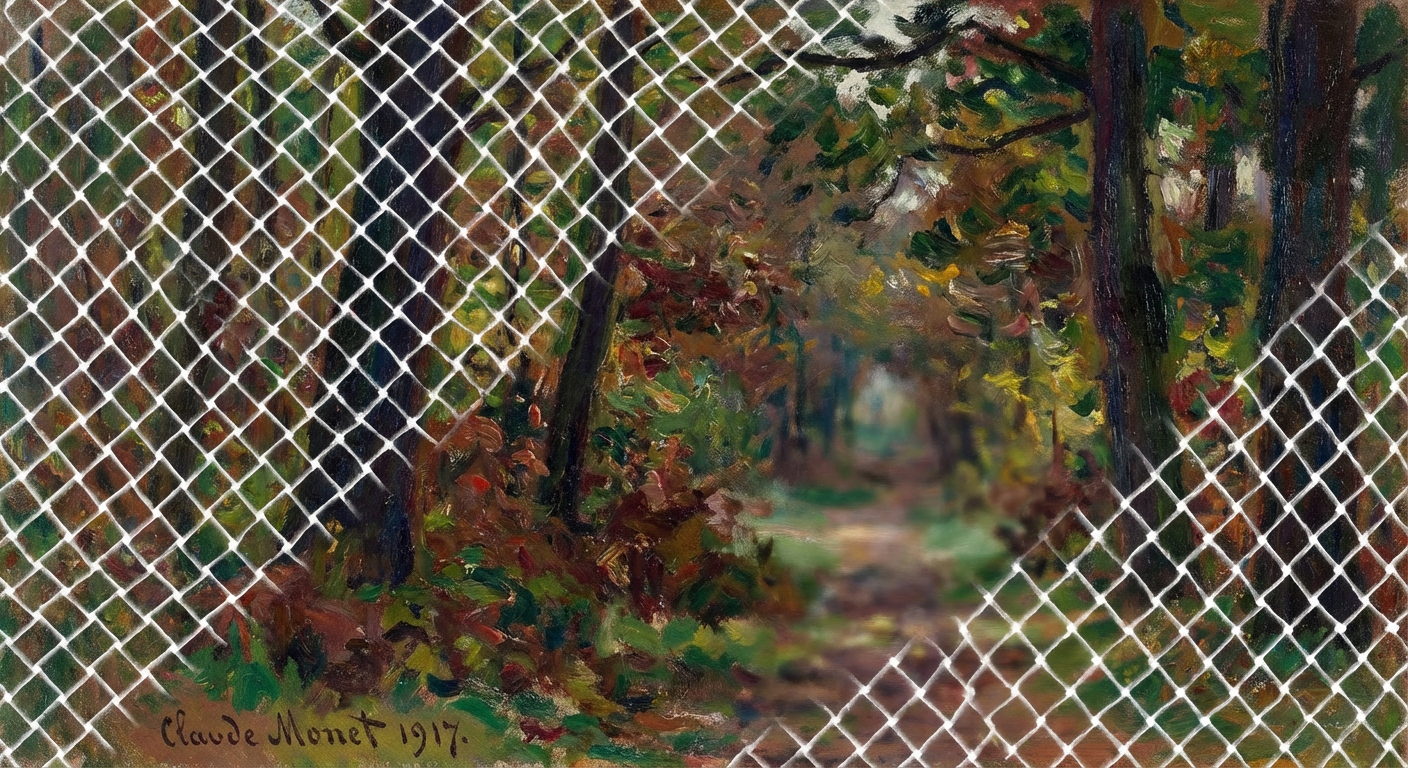} &
        \includegraphics[width=0.12\textwidth]{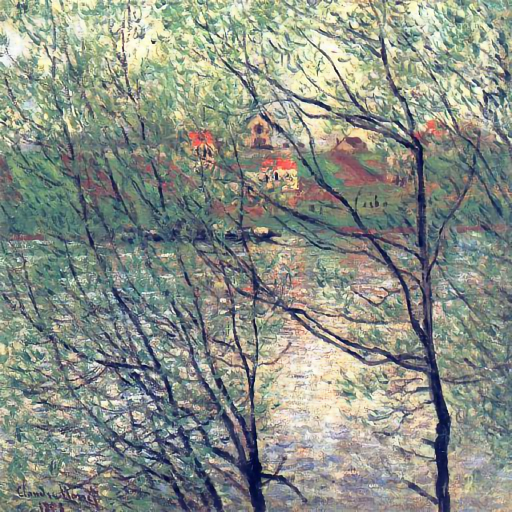} \\
        Low opacity & NB Pro & \\
        \includegraphics[width=0.12\textwidth]{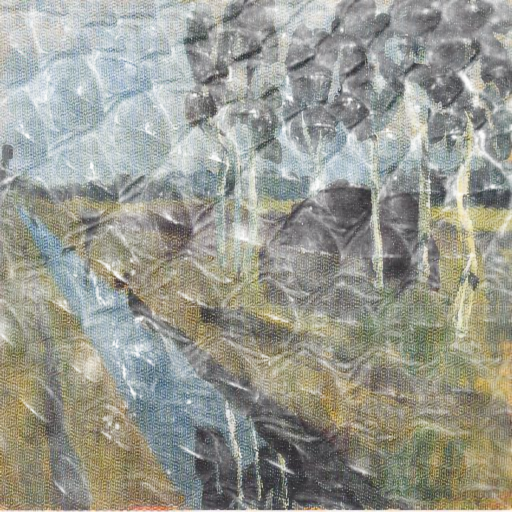} &
        \includegraphics[width=0.12\textwidth]{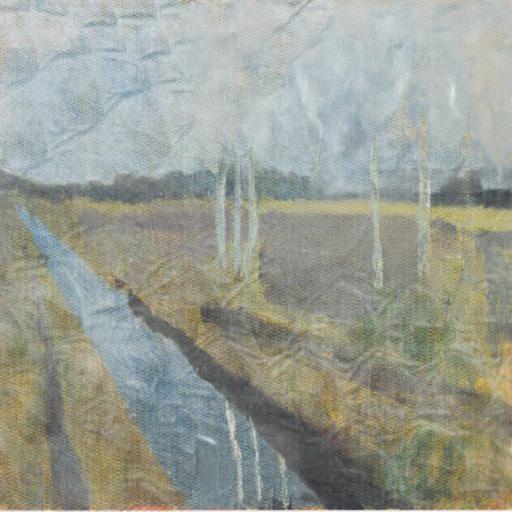} &
        \includegraphics[width=0.12\textwidth]{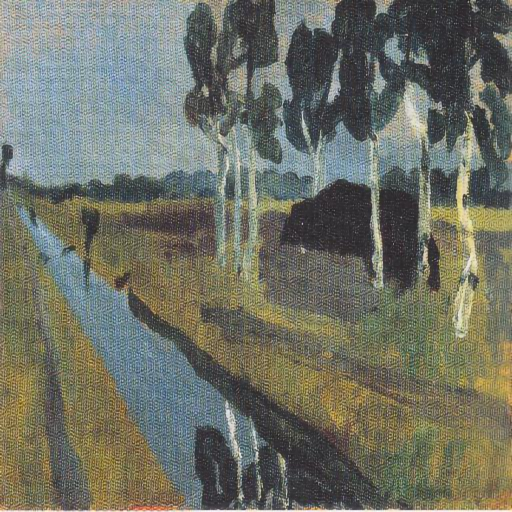} \\
        Low opacity & Step1x Edit & \\
        \includegraphics[width=0.12\textwidth]{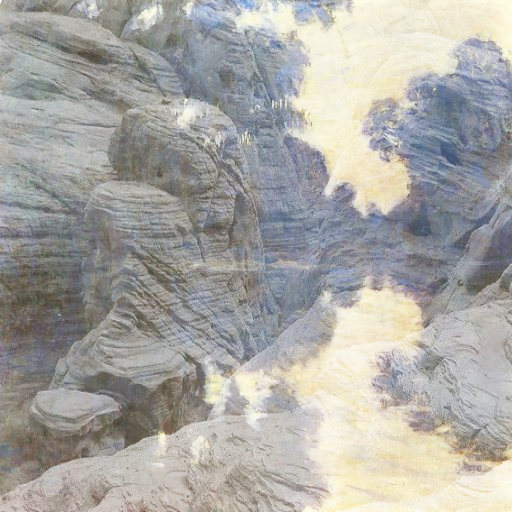} &
        \includegraphics[width=0.12\textwidth]{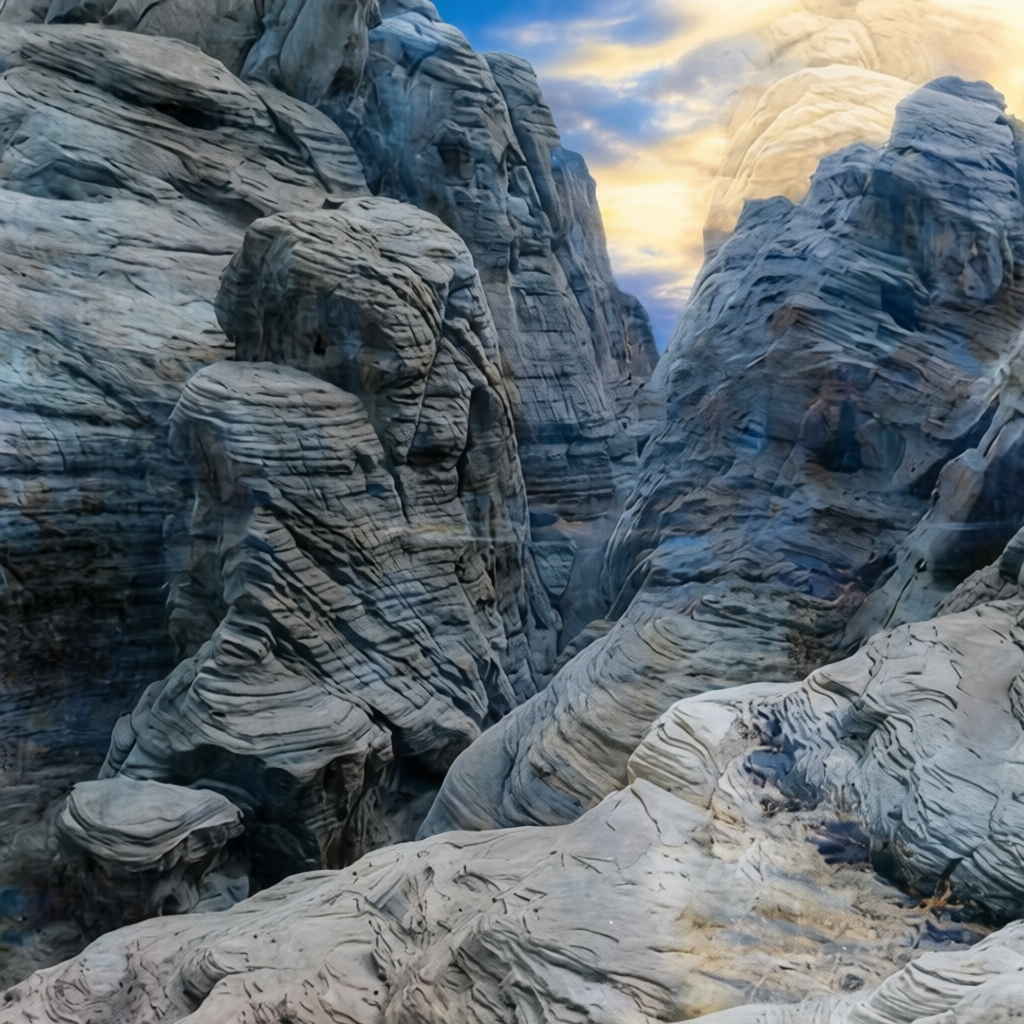} &
        \includegraphics[width=0.12\textwidth]{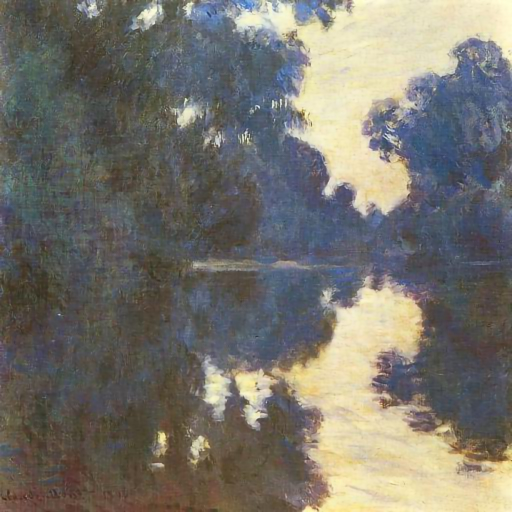} \\
        Medium opacity & Qwen Image Edit & \\
        \includegraphics[width=0.12\textwidth]{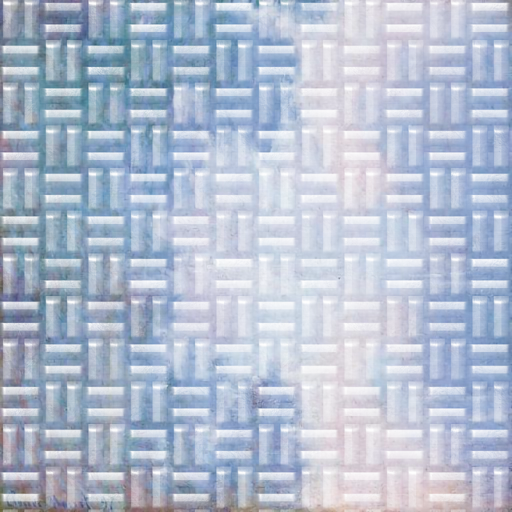} &
        \includegraphics[width=0.12\textwidth]{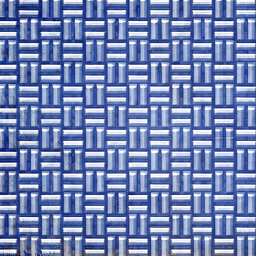} &
        \includegraphics[width=0.12\textwidth]{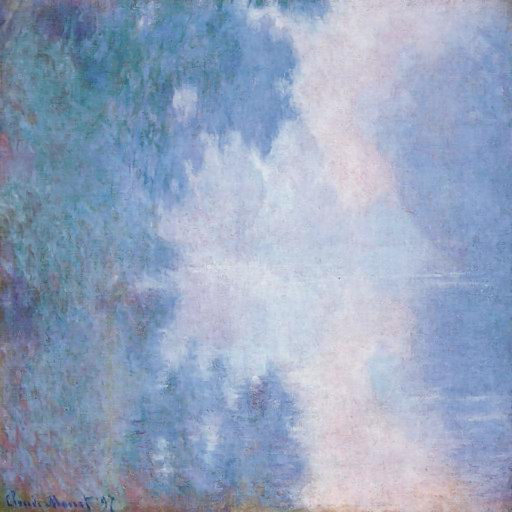} \\
        Low opacity & Flux 2 & \\

        \includegraphics[width=0.12\textwidth]{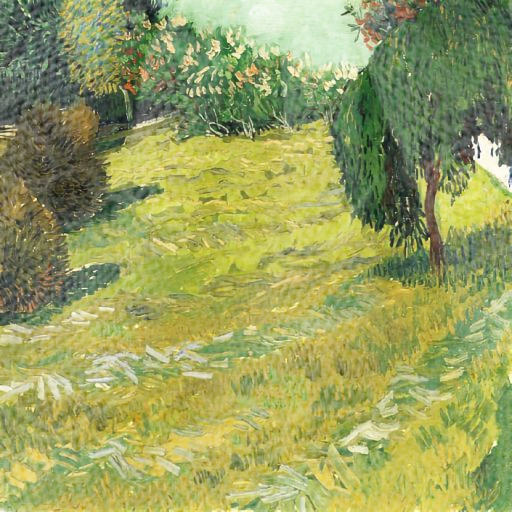} &
        \includegraphics[width=0.12\textwidth]{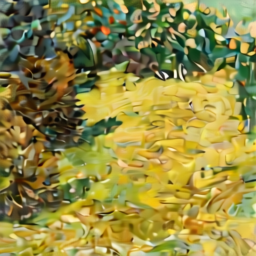} &
        \includegraphics[width=0.12\textwidth]{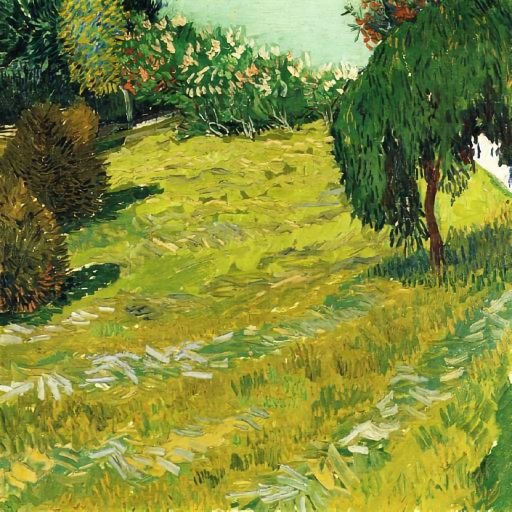} \\
        Low opacity & FireRed & \\
    \end{tabular}

    \caption{Hard degradation samples for different models  when utilizing the learned model priors for the non-explicit prompts (generic and explicit degradation-aware) under different levels of mask opacitys.}
    \label{fig:hard_samples}
\end{figure}

Fig. \ref{fig:realworld_samples} explores the transferability of our dataset masking to real-world degradations \cite{artworks_damaged_undamaged, heritage_paintings_restoration, xu2024comprehensive}. The image editing models such as Qwen Image Edit are prone to re-synthesizing, with the best restorations made with the closed NB Pro, which replaces the correct cracked and stained regions. Other image restoration mechanism having models such as AutoDIR and LanPaint(Qwen) preserve fidelity at the cost of leaving the damaged regions untouched. Step1x Edit gets a middle ground between these two extremes in inpainting or image restoration. AutoDIR and LanPaint(Qwen) have similar outputs, but the former incorporates more high-frequency details in the outputs.

The exploratory qualitative evaluation with real damage samples indicates that indicates that MDTD-Art's texture-mask degradations elicit the same restoration behaviours as authentic cracking, staining, and fading, suggesting similar failure behaviors in the restored images, albeit not including different coloring and mixing of underlying context information. The real-world damage samples are limited for any broadly generalizable insights, hence making this synthetic dataset an incomplete proxy until a collection comprehensively catalogs the different types of damage across media along with the respective restorations.

Fig. \ref{fig:hard_samples} illustrates the restorative performance of large image editing models on hard degradation masked images that result in the output image changing the identity of the underlying image and removing large parts of the image. Other models have their hard samples consisting of images retaining their degradation masks, since they are not the common types of degradations such as weather, compression, etc. Among generative image editing models, a recurring failure mode is the semantic dominance of the degradation mask itself. In NB Pro, grid-patterned and object-shaped masks overwhelm the underlying image semantics, even when the model demonstrably recognizes the mask structure, it cannot disentangle the mask from the scene to produce a faithful restoration, indicating a deviation in the semantic parsing and different task based instruction following (here, it would be the mask overlaid on the image is identified and treated differently under caption prompts and restoration prompts). FireRed Image Edit is specially trained for faces \cite{team2026firered}, which results in the face-patterned masks to hijack the restored image semantics. This is more common in Stable Diffusion 3, but the facial features in the result are not well defined and any small levels of facial details present in the underlying image itself are blurred. Flux 2, in contrast, produces more photorealistic images with vivid color palettes and strong shadows, a common characteristic in powerful image editing models such as GPT Image 1.5 and NB. They introduce extraneous scene elements such as miscounted objects and mountain geometry changes. Qwen Image Edit exhibits a yellow tone shift in the restored examples and any grayscale input examples are similarly toned without any explicit instructions regarding the colors. Step1x Edit is more conservative with its colors, blurring the results and occasionally leaving out some details in the restored image. BIRD has inconsistent performance for image restoration, missing out entire regions or retaining parts of the degradation mask due to it being non-binary, which is also why LanPaint does not work for inpainting. 
\section{Conclusion}
To overcome the limitations of existing universal image reconstruction that fails to address pixel-dependent and semantically ambiguous degradations, we introduce a new alpha texture mask dataset that provides a vital resource for future research on natural degradation of image media. We find that in the worst case scenario, the restored image loses the identity of the underlying image for image editing models, even if it retains the underlying color palette. We also observe that the image restoration models focus on retaining the identity of the underlying image over removing the alpha texture masks. By framing the task as a hybrid between restoration and inpainting, we exposed a regime that existing specialized restoration architectures are poorly equipped to handle, i.e. the case where  partial scene information must be leveraged alongside semantic understanding of the underlying content which are better utilized by image editing models for image restoration tasks. 

Our prompt ablation isolates variations of conceptual framing of the degradation operator (mask removal and counterfactual reconstruction) or inferred model priors. We find that semantic task disambiguation, rather than prompt complexity, is consistently correlated with restoration quality. For Nano Banana Pro and Flux 2, the observed benefit is greatest at high opacity. At high mask opacity, where little uncorrupted content remains, explicit task framing yields consistent gains for large editing models (e.g., Nano Banana Pro improves from 9.63 to 11.87 dB PSNR and Flux 2 from 8.57 to 11.01 dB), suggesting the richer semantic context partially compensates for the reduced visual evidence available to the model. As opacity decreases and models can increasingly rely on unmasked regions, the two prompt conditions restorations quality are similar for most models, though Nano Banana Pro and Step1x Edit 
retain a complex-prompt LPIPS advantage at medium opacity (0.35 vs.\ 0.36 and 0.36 vs.\ 0.38, 
respectively) and at low opacity (0.22 vs.\ 0.22 and 0.19 vs.\ 0.22, respectively). In contrast, restoration-specific models show minimal sensitivity to prompt type, as their behavior is dominated by input masks and learned degradation priors rather than textual conditioning. Together, these results position prompt design as a opacity-dependent source of restorative signal for language/instruction reliant models.

Finally, the strong performance of image editing models without restoration-specific supervision suggests that broad pretraining may contribute to the observed generalization, but controlled experiments are required to separate it from architecture, scale, and implementation effects. At the same time, persistent instability, hallucinations, and identity drift observed under adversarial mask conditions highlight that controllability and robustness remain open problems. Future work should explore soft masking strategies, degradation-aware prompt conditioning, and hybrid architectures that combine the semantic flexibility of editing models with the pixel-level fidelity constraints of discriminative restoration frameworks, particularly for the culturally significant but data-scarce domain of artistic heritage preservation.
\printcredits

\bibliographystyle{model1-num-names}

\bibliography{cas-refs}

@article{monga2025dairnet,
  title={DAIRNet: Degradation-aware All-in-one Image Restoration Network with cross-channel feature interaction},
  author={Monga, Amit and Nehete, Hemkant and Bollu, Tharun Kumar Reddy and Raman, Balasubramanian},
  journal={Journal of Visual Communication and Image Representation},
  pages={104659},
  year={2025},
  publisher={Elsevier}
}

@article{sah2025retrieval,
  title={Retrieval augmented generation for smart calorie estimation in complex food scenarios},
  author={Sah, Mayank and Suman, Saurya and Mathew, Jimson},
  journal={Journal of Visual Communication and Image Representation},
  pages={104632},
  year={2025},
  publisher={Elsevier}
}

@article{shamsuddin2022synthetic,
  title={From synthetic to natural—Single natural image dehazing deep networks using synthetic dataset domain randomization},
  author={Shamsuddin, Abdul Fathaah and Ragunathan, Krupasankari and PM, Deepak Raja Sekar and Sankaran, Praveen and others},
  journal={Journal of Visual Communication and Image Representation},
  volume={89},
  pages={103636},
  year={2022},
  publisher={Elsevier}
}

@article{fang2026region,
  title={Region-guided representation fusion and background consistency in mask-free image editing},
  author={Fang, YiXin and Zhang, Huaxiang and Ji, Hua and Dai, Xiaolong and Yao, Chunqiang},
  journal={Journal of Visual Communication and Image Representation},
  volume={119},
  pages={104852},
  year={2026},
  publisher={Elsevier}
}

@article{jin2025degradation,
  title={Degradation removal and detail restoration decomposition network for single image deraining},
  author={Jin, Jiyu and Qi, Xuanyu and Dong, Haobo and Guan, Qiyuan and Jin, Guiyue and Fan, Lei},
  journal={Journal of Visual Communication and Image Representation},
  pages={104520},
  year={2025},
  publisher={Elsevier}
}

@misc{diffusers_flux2_bnb4bit,
  title        = {{FLUX.2-dev-bnb-4bit}},
  author       = {{diffusers}},
  year         = {2025},
  howpublished = {Hugging Face},
  url          = {https://huggingface.co/diffusers/FLUX.2-dev-bnb-4bit}
}

@article{dettmers2023qlora,
  title={{QLoRA}: Efficient finetuning of quantized {LLM}s},
  author={Dettmers, Tim and Pagnoni, Artidoro and Holtzman, Ari and Zettlemoyer, Luke},
  journal={Advances in Neural Information Processing Systems},
  volume={36},
  year={2023}
}

@article{you2025enhancing,
  title={Enhancing aesthetic image generation with reinforcement learning guided prompt optimization in stable diffusion},
  author={You, Junyong and Lin, Yuan and Hu, Bin},
  journal={Journal of Visual Communication and Image Representation},
  pages={104641},
  year={2025},
  publisher={Elsevier}
}

@article{saleh2015large,
  title={Large-scale classification of fine-art paintings: Learning the right metric on the right feature},
  author={Saleh, Babak and Elgammal, Ahmed},
  journal={arXiv preprint arXiv:1505.00855},
  year={2015}
}

@misc{huggan_wikiart,
  title        = {{WikiArt Dataset}},
  author       = {{huggan}},
  year         = {2022},
  howpublished = {Hugging Face},
  url          = {https://huggingface.co/datasets/huggan/wikiart}
}

@misc{artworks_damaged_undamaged,
  author       = {{pes1ug22am047}},
  title        = {{DAMAGED\_AND\_UNDAMAGED\_ARTWORKS}},
  year         = {2024},
  howpublished = {Kaggle},
  note         = {Dataset},
  url          = {https://www.kaggle.com/datasets/pes1ug22am047/damaged-and-undamaged-artworks}
}

@misc{heritage_paintings_restoration,
  author       = {{colabsss}},
  title        = {{Digital Restoration of Heritage Paintings}},
  year         = {2024},
  howpublished = {Kaggle},
  note         = {Dataset},
  url          = {https://www.kaggle.com/datasets/colabsss/digital-restoration-of-heritage-paintings}
}

@InProceedings{cimpoi14describing,
	      Author    = {M. Cimpoi and S. Maji and I. Kokkinos and S. Mohamed and and A. Vedaldi},
	      Title     = {Describing Textures in the Wild},
	      Booktitle = {Proceedings of the {IEEE} Conf. on Computer Vision and Pattern Recognition ({CVPR})},
	      Year      = {2014}}

@inproceedings{jiang2024autodir,
  title={Autodir: Automatic all-in-one image restoration with latent diffusion},
  author={Jiang, Yitong and Zhang, Zhaoyang and Xue, Tianfan and Gu, Jinwei},
  booktitle={European Conference on Computer Vision},
  pages={340--359},
  year={2024},
  organization={Springer}
}

@article{lin2025harnessing,
  title={Harnessing diffusion-yielded score priors for image restoration},
  author={Lin, Xinqi and Yu, Fanghua and Hu, Jinfan and You, Zhiyuan and Shi, Wu and Ren, Jimmy S and Gu, Jinjin and Dong, Chao},
  journal={ACM Transactions on Graphics (TOG)},
  volume={44},
  number={6},
  pages={1--21},
  year={2025},
  publisher={ACM New York, NY, USA}
}

@article{liu2025step1x,
  title={Step1x-edit: A practical framework for general image editing},
  author={Liu, Shiyu and Han, Yucheng and Xing, Peng and Yin, Fukun and Wang, Rui and Cheng, Wei and Liao, Jiaqi and Wang, Yingming and Fu, Honghao and Han, Chunrui and others},
  journal={arXiv preprint arXiv:2504.17761},
  year={2025}
}

@misc{rombach2021highresolution,
      title={High-Resolution Image Synthesis with Latent Diffusion Models}, 
      author={Robin Rombach and Andreas Blattmann and Dominik Lorenz and Patrick Esser and Björn Ommer},
      year={2021},
      eprint={2112.10752},
      archivePrefix={arXiv},
      primaryClass={cs.CV}
}

@article{yilmaz2026edit2restore,
  title={Edit2Restore: Few-Shot Image Restoration via Parameter-Efficient Adaptation of Pre-trained Editing Models},
  author={Y{\i}lmaz, M Ak{\i}n and Bilican, Ahmet and Biner, Burak Can and Tekalp, A Murat},
  journal={arXiv preprint arXiv:2601.03391},
  year={2026}
}

@inproceedings{lin2024improving,
  title={Improving image restoration through removing degradations in textual representations},
  author={Lin, Jingbo and Zhang, Zhilu and Wei, Yuxiang and Ren, Dongwei and Jiang, Dongsheng and Tian, Qi and Zuo, Wangmeng},
  booktitle={Proceedings of the IEEE/CVF Conference on Computer Vision and Pattern Recognition},
  pages={2866--2878},
  year={2024}
}

@article{sun2024beyond,
  title={Beyond pixels: Text enhances generalization in real-world image restoration},
  author={Sun, Haoze and Li, Wenbo and Liu, Jiayue and Zhou, Kaiwen and Chen, Yongqiang and Guo, Yong and Li, Yanwei and Pei, Renjing and Peng, Long and Yang, Yujiu},
  journal={arXiv preprint arXiv:2412.00878},
  year={2024}
}

@article{bai2025textir,
  title={Textir: A simple framework for text-based editable image restoration},
  author={Bai, Yunpeng and Wang, Cairong and Xie, Shuzhao and Dong, Chao and Yuan, Chun and Wang, Zhi},
  journal={IEEE Transactions on Visualization and Computer Graphics},
  year={2025},
  publisher={IEEE}
}

@article{labs2025flux,
  title={FLUX. 1 Kontext: Flow Matching for In-Context Image Generation and Editing in Latent Space},
  author={Labs, Black Forest and Batifol, Stephen and Blattmann, Andreas and Boesel, Frederic and Consul, Saksham and Diagne, Cyril and Dockhorn, Tim and English, Jack and English, Zion and Esser, Patrick and others},
  journal={arXiv preprint arXiv:2506.15742},
  year={2025}
}

@article{team2023gemini,
  title={Gemini: a family of highly capable multimodal models},
  author={Team, Gemini and Anil, Rohan and Borgeaud, Sebastian and Alayrac, Jean-Baptiste and Yu, Jiahui and Soricut, Radu and Schalkwyk, Johan and Dai, Andrew M and Hauth, Anja and Millican, Katie and others},
  journal={arXiv preprint arXiv:2312.11805},
  year={2023}
}

@misc{openai2025gptimage15,
  author       = {OpenAI},
  title        = {Introducing GPT Image 1.5 — available today in the API and ChatGPT},
  howpublished = {OpenAI Developer Community announcement},
  year         = {2025},
  month        = dec,
  day          = {16},
  note         = {Released December 16, 2025},
  url          = {https://openai.com/index/new-chatgpt-images-is-here/}
}

@article{wu2025qwen,
  title={Qwen-image technical report},
  author={Wu, Chenfei and Li, Jiahao and Zhou, Jingren and Lin, Junyang and Gao, Kaiyuan and Yan, Kun and Yin, Sheng-ming and Bai, Shuai and Xu, Xiao and Chen, Yilei and others},
  journal={arXiv preprint arXiv:2508.02324},
  year={2025}
}

@article{team2026firered,
  title={Firered-image-edit-1.0 technical report},
  author={Team, Super Intelligence and Qiao, Changhao and Hui, Chao and Li, Chen and Wang, Cunzheng and Song, Dejia and Zhang, Jiale and Li, Jing and Xiang, Qiang and Wang, Runqi and others},
  journal={arXiv preprint arXiv:2602.13344},
  year={2026}
}

@inproceedings{chheda2025artinsight,
  title={ArtInsight: Enabling AI-Powered Artwork Engagement for Mixed Visual-Ability Families},
  author={Chheda-Kothary, Arnavi and Kanchi, Ritesh and Sanders, Chris and Xiao, Kevin and Sengupta, Aditya and Kneitmix, Melanie and Wobbrock, Jacob O and Froehlich, Jon E},
  booktitle={Proceedings of the 30th International Conference on Intelligent User Interfaces},
  pages={190--210},
  year={2025}
}

@misc{yang2023hq50k,
      title={HQ-50K: A Large-scale, High-quality Dataset for Image Restoration}, 
      author={Qinhong Yang and Dongdong Chen and Zhentao Tan and Qiankun Liu and Qi Chu and Jianmin Bao and Lu Yuan and Gang Hua and Nenghai Yu},
      year={2023},
      eprint={2306.05390},
      archivePrefix={arXiv},
      primaryClass={cs.CV}
}

@article{openai2025introducing,
  title={Introducing 4o image generation},
  author={OpenAI},
  year={2025},
  publisher={OpenAI, San Francisco CA, United States}
}

@inproceedings{lin2024diffbir,
  title={Diffbir: Toward blind image restoration with generative diffusion prior},
  author={Lin, Xinqi and He, Jingwen and Chen, Ziyan and Lyu, Zhaoyang and Dai, Bo and Yu, Fanghua and Qiao, Yu and Ouyang, Wanli and Dong, Chao},
  booktitle={European conference on computer vision},
  pages={430--448},
  year={2024},
  organization={Springer}
}

@article{liu2024adaptbir,
  title={AdaptBIR: Adaptive Blind Image Restoration with latent diffusion prior for higher fidelity},
  author={Liu, Yingqi and He, Jingwen and Liu, Yihao and Lin, Xinqi and Yu, Fanghua and Hu, Jinfan and Qiao, Yu and Dong, Chao},
  journal={Pattern Recognition},
  volume={155},
  pages={110659},
  year={2024},
  publisher={Elsevier}
}

@inproceedings{yu2024scaling,
  title={Scaling up to excellence: Practicing model scaling for photo-realistic image restoration in the wild},
  author={Yu, Fanghua and Gu, Jinjin and Li, Zheyuan and Hu, Jinfan and Kong, Xiangtao and Wang, Xintao and He, Jingwen and Qiao, Yu and Dong, Chao},
  booktitle={Proceedings of the IEEE/CVF conference on computer vision and pattern recognition},
  pages={25669--25680},
  year={2024}
}

@article{tang2025degradation,
  title={Degradation-aware residual-conditioned optimal transport for unified image restoration},
  author={Tang, Xiaole and Gu, Xiang and He, Xiaoyi and Hu, Xin and Sun, Jian},
  journal={IEEE Transactions on Pattern Analysis and Machine Intelligence},
  year={2025},
  publisher={IEEE}
}

@inproceedings{chen2024comparative,
  title={A comparative study of image restoration networks for general backbone network design},
  author={Chen, Xiangyu and Li, Zheyuan and Pu, Yuandong and Liu, Yihao and Zhou, Jiantao and Qiao, Yu and Dong, Chao},
  booktitle={European Conference on Computer Vision},
  pages={74--91},
  year={2024},
  organization={Springer}
}

@inproceedings{guo2024onerestore,
  title={Onerestore: A universal restoration framework for composite degradation},
  author={Guo, Yu and Gao, Yuan and Lu, Yuxu and Zhu, Huilin and Liu, Ryan Wen and He, Shengfeng},
  booktitle={European conference on computer vision},
  pages={255--272},
  year={2024},
  organization={Springer}
}

@article{eteke2026bir,
  title={BIR-Adapter: A parameter-efficient diffusion adapter for blind image restoration},
  author={Eteke, Cem and Griessel, Alexander and Kellerer, Wolfgang and Steinbach, Eckehard},
  journal={Pattern Recognition},
  pages={113824},
  year={2026},
  publisher={Elsevier}
}

@article{zheng2025lanpaint,
  title={LanPaint: Training-Free Diffusion Inpainting with Asymptotically Exact and Fast Conditional Sampling},
  author={Zheng, Candi and Lan, Yuan and Wang, Yang},
  journal={arXiv preprint arXiv:2502.03491},
  year={2025}
}

@article{wang2004image,
  title={Image quality assessment: from error visibility to structural similarity},
  author={Wang, Zhou and Bovik, Alan C and Sheikh, Hamid R and Simoncelli, Eero P},
  journal={IEEE transactions on image processing},
  volume={13},
  number={4},
  pages={600--612},
  year={2004},
  publisher={IEEE}
}

@inproceedings{zhang2018unreasonable,
  title={The unreasonable effectiveness of deep features as a perceptual metric},
  author={Zhang, Richard and Isola, Phillip and Efros, Alexei A and Shechtman, Eli and Wang, Oliver},
  booktitle={Proceedings of the IEEE conference on computer vision and pattern recognition},
  pages={586--595},
  year={2018}
}

@article{Luo-Rephotography-2021,
  author    = {Luo, Xuan and Zhang, Xuaner and Yoo, Paul and Martin-Brualla, Ricardo and Lawrence, Jason and Seitz, Steven M.},
  title     = {Time-Travel Rephotography},
  journal = {ACM Transactions on Graphics (Proceedings of ACM SIGGRAPH Asia 2021)},
  publisher = {ACM New York, NY, USA},
  volume = {40},
  number = {6},
  articleno = {213},
  doi = {https://doi.org/10.1145/3478513.3480485},
  year = {2021},
  month = {12}
}

@inproceedings{wan2020bringing,
title={Bringing Old Photos Back to Life},
author={Wan, Ziyu and Zhang, Bo and Chen, Dongdong and Zhang, Pan and Chen, Dong and Liao, Jing and Wen, Fang},
booktitle={Proceedings of the IEEE/CVF Conference on Computer Vision and Pattern Recognition},
pages={2747--2757},
year={2020}
}

@article{zuo2025nano,
  title={Is Nano Banana Pro a Low-Level Vision All-Rounder? A Comprehensive Evaluation on 14 Tasks and 40 Datasets},
  author={Zuo, Jialong and Deng, Haoyou and Zhou, Hanyu and Zhu, Jiaxin and Zhang, Yicheng and Zhang, Yiwei and Yan, Yongxin and Huang, Kaixing and Chen, Weisen and Deng, Yongtai and others},
  journal={arXiv preprint arXiv:2512.15110},
  year={2025}
}

@article{xu2024comprehensive,
  title={A comprehensive dataset for digital restoration of Dunhuang murals},
  author={Xu, Zishan and Yang, Yuqing and Fang, Qianzhen and Chen, Wei and Xu, Tingting and Liu, Jueting and Wang, Zehua},
  journal={Scientific Data},
  volume={11},
  number={1},
  pages={955},
  year={2024},
  publisher={Nature Publishing Group UK London}
}

@article{garcia2025artinsight,
  title={ArtInsight: A detailed dataset for detecting deterioration in easel paintings},
  author={Garcia-Moreno, Francisco M and Rodr{\'\i}guez-Sim{\'o}n, Luis Rodrigo and Hurtado-Torres, Mar{\'\i}a Visitaci{\'o}n and others},
  journal={Data in Brief},
  volume={61},
  pages={111811},
  year={2025},
  publisher={Elsevier}
}

@inproceedings{shah2020performance,
  title={A performance analysis of deep convolutional neural networks using Kuzushiji character recognition},
  author={Shah, Harshil and Manjula, V},
  booktitle={2020 International Conference on Decision Aid Sciences and Application (DASA)},
  pages={1068--1071},
  year={2020},
  organization={IEEE}
}

@article{yang2026realrestorer,
  title={RealRestorer: Towards Generalizable Real-World Image Restoration with Large-Scale Image Editing Models},
  author={Yang, Yufeng and Zeng, Xianfang and Jiang, Zhangqi and Yin, Fukun and Liu, Jianzhuang and Cheng, Wei and Liu, Shiyu and Peng, Yuqi and YU, Gang and Chen, Shifeng and others},
  journal={arXiv preprint arXiv:2603.25502},
  year={2026}
}

@article{yang2023hq,
  title={Hq-50k: A large-scale, high-quality dataset for image restoration},
  author={Yang, Qinhong and Chen, Dongdong and Tan, Zhentao and Liu, Qiankun and Chu, Qi and Bao, Jianmin and Yuan, Lu and Hua, Gang and Yu, Nenghai},
  journal={arXiv preprint arXiv:2306.05390},
  year={2023}
}

@inproceedings{sun2026can,
  title={Can Nano Banana 2 Replace Traditional Image Restoration Models? An Evaluation of Its Performance on Image Restoration Tasks},
  author={Sun, Weixiong and Yin, Xiang and Dong, Chao},
  booktitle={Proceedings of the IEEE/CVF Conference on Computer Vision and Pattern Recognition},
  pages={4581--4590},
  year={2026}
}

@inproceedings{yin2026far,
  title={How far have we gone in Generative Image Restoration? A study on its capability, limitations and evaluation practices},
  author={Yin, Xiang and Hu, Jinfan and You, Zhiyuan and Yan, Kainan and Tang, Yu and Dong, Chao and Gu, Jinjin},
  booktitle={Proceedings of the IEEE/CVF Conference on Computer Vision and Pattern Recognition},
  pages={4909--4919},
  year={2026}
}

@inproceedings{yang2023pasd,
    title={Pixel-Aware Stable Diffusion for Realistic Image Super-Resolution and Personalized Stylization},
    author={Tao Yang and Rongyuan Wu and Peiran Ren and Xuansong Xie and and Lei Zhang},
    booktitle={The European Conference on Computer Vision (ECCV) 2024},
    year={2023}
}

@inproceedings{guo2025mambair,
  title={MambaIR: A simple baseline for image restoration with state-space model},
  author={Guo, Hang and Li, Jinmin and Dai, Tao and Ouyang, Zhihao and Ren, Xudong and Xia, Shu-Tao},
  booktitle={European Conference on Computer Vision},
  pages={222--241},
  year={2024},
  organization={Springer}
}

@article{chihaoui2024blind,
   title={Blind image restoration via fast diffusion inversion},
   author={Chihaoui, Hamadi and Lemkhenter, Abdelhak and Favaro, Paolo},
   journal={Advances in Neural Information Processing Systems},
   volume={37},
   pages={34513--34532},
   year={2024}
 }

@article{cai2024hierarchical,
  title={Hierarchical damage correlations for old photo restoration},
  author={Cai, Weiwei and Xu, Xuemiao and Xu, Jiajia and Zhang, Huaidong and Yang, Haoxin and Zhang, Kun and He, Shengfeng},
  journal={Information Fusion},
  volume={107},
  pages={102340},
  year={2024},
  publisher={Elsevier}
}

@article{o2023limitations,
  title={Limitations and possibilities of digital restoration techniques using generative AI tools: Reconstituting Antoine Fran{\c{c}}ois Callet’s Achilles Dragging Hector’s Body Past the Walls of Troy},
  author={O'Brien, Charles and Hutson, James and Olsen, Trent and Ratican, Jay},
  journal={Arts \& Communication},
  year={2023}
}

@inproceedings{huang2024learning,
  title={Learning disentangled identifiers for action-customized text-to-image generation},
  author={Huang, Siteng and Gong, Biao and Feng, Yutong and Chen, Xi and Fu, Yuqian and Liu, Yu and Wang, Donglin},
  booktitle={Proceedings of the IEEE/CVF Conference on Computer Vision and Pattern Recognition},
  pages={7797--7806},
  year={2024}
}

@article{vinker2023concept,
  title={Concept decomposition for visual exploration and inspiration},
  author={Vinker, Yael and Voynov, Andrey and Cohen-Or, Daniel and Shamir, Ariel},
  journal={ACM Transactions on Graphics (TOG)},
  volume={42},
  number={6},
  pages={1--13},
  year={2023},
  publisher={ACM New York, NY, USA}
}

@inproceedings{avrahami2023break,
   title={Break-a-scene: Extracting multiple concepts from a single image},
   author={Avrahami, Omri and Aberman, Kfir and Fried, Ohad and Cohen-Or, Daniel and Lischinski, Dani},
   booktitle={SIGGRAPH Asia 2023 Conference Papers},
   pages={1--12},
   year={2023}
 }

@inproceedings{safaee2024clic,
  title={Clic: Concept learning in context},
  author={Safaee, Mehdi and Mikaeili, Aryan and Patashnik, Or and Cohen-Or, Daniel and Mahdavi-Amiri, Ali},
  booktitle={Proceedings of the IEEE/CVF Conference on Computer Vision and Pattern Recognition},
  pages={6924--6933},
  year={2024}
}

@inproceedings{wang2025towards,
  title={Towards Enhanced Image Inpainting: Mitigating Unwanted Object Insertion and Preserving Color Consistency.},
  author={Wang, Yikai and Cao, Chenjie and Yu, Junqiu and Fan, Ke and Xue, Xiangyang and Fu, Yanwei},
  booktitle={Proceedings of the IEEE/CVF conference on computer vision and pattern recognition},
  year={2025}
}

@article{vijendran25artificial,
 author={Vijendran, Mridula and Deng, Jingjing and Chen, Shuang and Ho, Edmond S. L. and Shum, Hubert P. H.},
 journal={Artificial Intelligence Review},
 title={Artificial Intelligence for Geometry-Based Feature Extraction, Analysis and Synthesis in Artistic Images: A Survey},
 year={2024},
 volume={58},
 number={2},
 pages={64},
 numpages={47},
 doi={10.1007/s10462-024-11051-3},
 issn={1573-7462},
 publisher={Springer},
}

@article{chen24hint,
 author={Chen, Shuang and Atapour-Abarghouei, Amir and Shum, Hubert P. H.},
 journal={IEEE Transactions on Multimedia},
 title={HINT: High-quality INpainting Transformer with Mask-Aware Encoding and Enhanced Attention},
 year={2024},
 volume={26},
 pages={7649--7660},
 numpages={12},
 doi={10.1109/TMM.2024.3369897},
 issn={1520-9210 },
 publisher={IEEE},
}

@article{chang26design,
 author={Chang, Ziyi and Koulieris, George Alex and Chang, Hyung Jin and Shum, Hubert P. H.},
 journal={Pattern Recognition},
 title={On the Design Fundamentals of Diffusion Models: A Survey},
 year={2026},
 pages={111934},
 doi={10.1016/j.patcog.2025.111934},
 issn={0031-3203},
 publisher={Elsevier},
}



\end{document}